\documentclass[11pt,a4paper,table]{article}
\usepackage[hyperref]{AACL-IJCNLP2020}
\usepackage{times}
\usepackage{latexsym}
\usepackage{enumitem}

\usepackage{microtype}

\usepackage{booktabs} % import booktabs for tabulars rulers
\usepackage{tabularx}
\usepackage{float} % enable H option to prevent tables from exceeding the linewidth
\usepackage{graphicx} % to include graphics
\usepackage{multirow} % provide the functionality of grouping multiple rows
\usepackage{diagbox} % to divide a rectangle by a diagonal line in the result tables
\usepackage{pifont}% Zapf-Dingbats Symbole
\usepackage{subfigure}
\usepackage{amsmath}
\usepackage{amssymb}
\usepackage{fp-basic} % decimal if else
\usepackage{pgf}
\usepackage{collcell}
\usepackage[acronym]{glossaries}
\usepackage{xcolor}
\glsdisablehyper

\usepackage{siunitx}
\usepackage{pgfplotstable}
\pgfplotsset{compat=1.16} 

\usepackage[utf8x]{inputenc}
\usepackage{tipa}

\newcommand\customfont[1]{{\usefont{T1}{Noto Sans}{m}{n} #1 }}

\DeclareUnicodeCharacter{605}{\textrhookrevepsilon{}}
\DeclareUnicodeCharacter{619}{\textltilde{}}
\DeclareUnicodeCharacter{7428}{c}
\DeclareUnicodeCharacter{639}{\textfishhookr{}}
\DeclareUnicodeCharacter{7597}{\textturnmrleg{}}
\DeclareUnicodeCharacter{671}{\textscl{}}

\usepackage[normalem]{ulem}

\newcommand*{\thead}[1]{\bfseries #1} % define a command for table header

\newcommand{\cmark}{\ding{51}}%
\newcommand{\xmark}{\ding{55}}%

\newacronym{tsne}{t-SNE}{t-distributed stochastic neighbor embedding}
\newacronym{ai}{AI}{Artifical Intelligence}
\newacronym{nlp}{NLP}{Natural Language Processing}
\newacronym{cv}{CV}{Computer Vision}
\newacronym{sota}{SOTA}{state-of-the-art}
\newacronym{mlm}{MLM}{Masked Language Model}
\newacronym{plm}{PLM}{Perturbed Language Model}
\newacronym{at}{AT}{Adversarial Training}
\newacronym{sem}{SEM}{Synonym Encoding Method}
\newacronym{scpn}{SCPN}{Syntactically Controlled Paraphrase Network}
\newacronym{at11}{AT 1-1}{Adversarial Training 1-1}
\newacronym{atloo}{AT LOO}{Adversarial Training Leave-One-Out}
\newacronym{pp}{PP}{Perturbed Pretraining}
\newacronym{ml}{ML}{machine learning}
\newacronym{mt}{MT}{machine translation}
\newacronym{pos}{POS}{Part-of-Speech}
\newacronym{snli}{SNLI}{Stanford Natural Language Inference}
\newacronym{tc}{TC}{Toxic Comment}
\newacronym{nli}{NLI}{Natural Language Inference}
\newacronym{lmh}{LMH}{low-mid-high}
\newacronym{clmh}{CLMH}{clean-low-mid-high}
\newacronym{cnn}{CNN}{Convolutional Neural Network}
\newacronym{rnn}{RNN}{Recurrent Neutral Network}
\newacronym{lstm}{LSTM}{Long Short-Term Memory}
\newacronym{gru}{GRU}{Gated Recurrent Units}
\newacronym{bilstm}{BiLSTM}{Bi-directional LSTM}
\newacronym{mlp}{MLP}{Multilayer Perceptron}
\newacronym{gan}{GAN}{Generative Adversarial Network}
\newacronym{dnn}{DNN}{Deep Neural Network}
\newacronym{cae}{CAE}{Convolutional Autoencoder}
\newacronym{vae}{VAE}{Variational Autoencoder}
\newacronym{vaegan}{VAEGAN}{Variational Autoencoder Generative Adversarial Network}
\newacronym{g2p}{G2P}{grapheme-to-phoneme}
\newacronym{p2g}{P2G}{phoneme-to-grapheme}

\newcommand*{\MinNumber}{0}%
\newcommand*{\MaxNumber}{1}%

\definecolor{Red}{rgb}{1,0,0}
\definecolor{Blue}{rgb}{0,0,1}

\newcommand{\ApplyGradient}[1]{%
    \FPiflt{#1}{0}
        \pgfmathsetmacro{\PercentColor}{-100.0*(#1-\MinNumber)/(\MaxNumber-\MinNumber)}%
        \edef\x{\noexpand\cellcolor{Red!\PercentColor!Blue}}\x\textcolor{white}{#1}
    \else
        \pgfmathsetmacro{\PercentColor}{100.0*(#1-\MinNumber)/(\MaxNumber-\MinNumber)}%
        \edef\x{\noexpand\cellcolor{Blue!\PercentColor!Red}}\x\textcolor{white}{#1}%
    \fi
}

\newcommand*{\MinNumberHundred}{0}%
\newcommand*{\MaxNumberHundred}{100}

\newcommand{\ApplyGradientTwo}[1]{%
    \FPiflt{#1}{0}
        \pgfmathsetmacro{\PercentColor}{-100.0*(#1-\MinNumberHundred)/(\MaxNumberHundred-\MinNumberHundred)}%
        \edef\x{\noexpand\cellcolor{Red!\PercentColor!Blue}}\x\textcolor{white}{#1}
    \else
        \pgfmathsetmacro{\PercentColor}{100.0*(#1-\MinNumberHundred)/(\MaxNumberHundred-\MinNumberHundred)}%
        \edef\x{\noexpand\cellcolor{Blue!\PercentColor!Red}}\x\textcolor{white}{#1}%
    \fi
}

\newcolumntype{R}{>{\collectcell\ApplyGradient}{r}<{\endcollectcell}}
\newcolumntype{K}{>{\collectcell\ApplyGradientTwo}{r}<{\endcollectcell}}

\definecolor{gold}{RGB}{255,215,0}
\definecolor{silver}{RGB}{192,192,192}
\definecolor{bronze}{RGB}{140,120,83}

\aclfinalcopy % Uncomment this line for the final submission
\newcommand{\checkme}[1]{\textcolor{black}{#1}}
\title{From Hero to Zéroe: \\
A Benchmark of Low-Level Adversarial Attacks}

\author{Steffen Eger and Yannik Benz\\
    Computer Science Department\\
    Technische Universit\"at Darmstadt, Germany \\
     \texttt{{eger}@aiphes.tu-darmstadt.de}, \texttt{yannik.benz@stud.tu-darmstadt.de}\\ %\texttt{\{daxenberger,gurevych\}@ukp.informatik.tu-darmstadt.de} \\
 }

\date{}

\begin{document}
\maketitle
\begin{abstract}
Adversarial attacks are label-preserving modifications to inputs of machine learning classifiers designed to fool machines but not humans. \acrfull{nlp} has mostly focused on high-level attack scenarios such as paraphrasing input texts. We argue that these are less realistic in typical application scenarios such as in social media, and instead focus on \emph{low-level} attacks on the character-level. Guided by human cognitive abilities and human robustness, we propose the first large-scale catalogue and benchmark of low-level adversarial attacks, which we dub \emph{Zéroe}, encompassing nine different attack modes including visual and phonetic adversaries. We show that \emph{RoBERTa}, \acrshort{nlp}'s current workhorse, fails on our attacks. Our dataset provides a benchmark for testing robustness of future more human-like \acrshort{nlp} models.  
%In NLP, deep-learning models have been found to be susceptible to \textit{adversarial attacks}. This work catalogues ten different low-level adversarial attacks: full-shuffle, inner-shuffle, intrude, disemvowel, truncate, segmentation, keyboard-typos, natural-noise, phonetic, and visual. % maximum of 200 words
\end{abstract}

\section{Introduction}
\label{sec:introduction}
%
%
%
%\todo[inline]{
%\begin{enumerate}
   % \item fix acronyms to be CAPS
    %\begin{enumerate}
    %    \item check acronym in abstract
    %\end{enumerate}
    %\item For each experiment, training hyper parameters (such as batch sizes, epochs, and learning rates) should be reported. Or, at least, the authors should mention where such information is available.
    %\item Replace \S by Sec as recommended by reviewer to meet sytle guidelines
    %\item insert github link as footnote to the respective place \footnote{https://github.com/yannikbenz/robustness-of-deep-learning-models-to-low-level-adversarial-attacks} and make repository public and maybe further cleanup code
    %\item upload the datasets
%\end{enumerate}
%}
%
%
%
Adversarial examples are label-preserving modifications to inputs of machine learning architectures. Their typical characteristic is that they cause little damage to humans %text processing 
but may maximally affect classifier performance, exposing their weaknesses and outlining the differences between human and machine text processing \citep{Szegedy2014Intriguing,Goodfellow2014Explaining,eger2019text}. 

While in computer vision, pixel-level attacks, which go unnoticed by humans, may lead to catastrophic failure, attacks in NLP are more challenging. Some attacks in NLP replace individual words by synonyms or hyponyms \citep{Alzantot2018Generating} or paraphrase  whole sentences \citep{Ribeiro2018Semantically}. However, such \emph{high-level} attacks are not only more difficult to compute (requiring available resources such as dictionaries or word embeddings) but they are also implausible in real-world scenarios such as spamming or posting in \emph{social media}, as users would need to know the training data and/or the inner workings of the machine learning models in order to identify candidate substitutions (or have unrestrained access to model predictions). In contrast, such users would typically use \emph{low-level} attacks on characters, such as inserting placeholder symbols (e.g., underscores), mistyping words (e.g., \emph{Hilter} for \emph{Hitler}), or using phonetically similar sounding words \citep{tagg2011wot} to fool online detection models. To identify plausible such attack scenarios, human perceptual abilities play a decisive role. For instance, humans are guided by their senses,  making them robust to, e.g., visual and phonetic attacks. Other scenarios to which humans have been shown robust include the removal of vowels from words or the shuffling of characters while keeping the initial and final letters fixed (see Section \ref{sec:related_work}). However, the varieties in which text can be perturbed is certainly far from infinite, as (ordinary) humans, with all their cognitive constraints, still need to be able to decipher the text messages.

\begin{table}
    \centering
    \begin{tabularx}{\linewidth}{lX}
    \toprule
    % begin table header
    \textbf{Attacker} & \textbf{Sentence} \\
    \midrule
    % end table header
    % begin table body
        inner-shuffle & Aadrreavsil aacttks are hmarsels. \\ \hline
        full-shuffle & idaAasvrler tstkaac are harmless. \\ \hline
        intrude & A d v e r sar ial at:ta:ck:s are h\}ar\}m\}less. \\ \hline
        disemvowel & dvrsrl ttcks r hrmlss. \\ \hline
        truncate & Adversaria attack are harmles. \\ \hline
        segment & Adversarial attacksare harmless. \\ \hline
        typo & Adverssrial attaxks are harmless. \\ \hline
        natural noise & Adversarial attacs rae harmless. \\ \hline
        phonetic & %Adveraciory attacks are harmless. 
        Advorcariel attaks are harmless. 
        \\ 
        \hline
        visual & \customfont{Ǟḋ\textupsilon{}ɝrsarīaɫ aṯtaᴄḳṣ āɿe hǟrᶭʟ\~{e}ṣ.} \\
    % end table body
    \bottomrule
    \end{tabularx}
    \caption{Ten different modifications of the sentence ``Adversarial attacks are harmless.''}
    \label{table:my_attacks}
\end{table}

In this work, we %aim %at providing 
provide 
the first large-scale   
catalogue for \emph{low-level} (orthographic) attack scenarios. Our search is motivated by insights into human cognitive limitations and constraints and encompasses nine different attack modes (some of which are overlapping); cf.\ Table~\ref{table:my_attacks}.  
We then examine the robustness of \emph{RoBERTa} \citep{liu2019roberta} to our attacks, finding that its performance can %almost maximally 
sometimes be severely decreased for %a majority of 
our selection of attackers (up to the random guessing baseline); hence we call our benchmark \emph{Zéroe}. 
The reason may be that our noises are not always natural, in the sense of having high support in large datasets such as CommonCrawl or Wikipedia, but they are still within the limits of cognitive abilities of ordinary humans. Finally, we show that under realistic conditions, standard adversarial training can restore \emph{RoBERTa}'s performance only to a limited degree.\footnote{Code and data are provided at \url{https://github.com/yannikbenz/zeroe}.}

% fine

\section{Related Work}
\label{sec:related_work}
%\begin{itemize}
%    \item this chapter introduces scientific research
%\end{itemize}

%\subsection{Adversarial Attacks}

%Adversarial attacks %are the process of inserting modified %inputs into a deep-neural network in order 
%aim to fool machine learning systems into wrong decisions  \citep{zhang2019adversarial}, while at the same time, the original meaning should still be understandable by humans. %
%The attacks can be separated 
We classify adversarial attacks into \emph{high-} and \emph{low-level} attacks.\footnote{As one reviewer points out, a conceptual difference between high- and low-level attacks is that low-level attacks (as we define them) oftentimes induce linguistically corrupt text which can still be understood by humans, while high-level attacks operate in a noise-free environment to show the brittleness of systems even under `normal' circumstances.} %
%Under low-level attacks, we subsume attacks that [...] 
%Under high-level attacks, we subsume attacks that [...]
%In the following character-based attacks will be mentioned as \textbf{low-level} attacks, i.e., the smallest textual unit that will be changed is the character. Modifications can be made to single- or multiple characters. 

\paragraph{Attack Scenarios.}
There are a variety of works %that fool deep learning models on the character-level. 
that introduce \textbf{low-level} orthographic  attacks.\footnote{Low-level adversarial attacks are in part examined by approaches to handle noisy user-generated text \citep{baldwin2015shared}, with one difference being that attacks are often malicious in nature and may thus come in different forms.} 
\citet{Ebrahimi2018Hotflip} %and \citet{Liang2017} 
%calculated 
%determine 
%the most damaging character addition, removal or flip for a character-level model and evaluated it on textual entailment. %task.
trick a character-level neural text classification model by flipping the characters which cause most damage. Their approach is white-box, i.e., assumes access to the attack model's parameters. 
\citet{eger2019text} %introduced an approach to 
exchange characters with similar looking ones %with each other 
%and showed its effectiveness on different character-, word- and sentence-based \acrshort{nlp} tasks. 
and show that humans are robust to such visual perturbations, while machines may suffer severe performance drops. 
\citet{Belinkov2017synthetic} %introduce keyboard-typos, 
exchange %in which 
adjacent letters on the keyboard %are exchanged 
with each other (keyboard typos) and introduce natural noise based on human typing errors extracted from different Wikipedia edit histories, %are applied to words, 
as well as %letter-swaps---multi-character perturbations in which two or more letters are swapped - and used them to show the brittleness of machine translation. 
letter swaps. They use this natural and synthetic noise to show the brittleness of \acrfull{mt} systems, which contrasts with corresponding human robustness. 
\citet{Ebrahimi2018OnAdversarial} also fool \acrshort{mt} systems with character-level modifications. \citet{tan-etal-2020-morphin} attack words by replacing them with morphological variants, which also mostly results in orthographic attacks (in English).

%introduce an attacker which aims to remove specific words of a machine translation or to change it entirely. They accomplish this by perturbing words in the input via replacing one or multiple characters with others, or by replacing full words with random letter sequences.%

\textbf{High-level} attacks require a deeper understanding of the meaning and the syntactical structure of the sentence. %That is because the modification of bigger textual units can easily mess up the correct grammatical structure or change the meaning of the whole sentence.
%Different word- and sentence-based methods to generate such examples have been investigated.
\citet{Jin2019IsBERT} %implemented an attacker to 
generate semantically similar and syntactically correct adversarial examples by replacing words with suitable %appropriate 
synonyms. %They accomplished this by identifying important words for the model via a preprocessing step and changed their inputs accordingly. 
%They evaluated multiple \acrshort{sota} models for text classification and entailment against their attacks.
\citet{Hosseini2017Deceiving} and \citet{Rodriguez2018Shielding} attack %the a 
toxic detection systems by obfuscation, i.e., misspelling of the abusive words (a low-level attack), %\todo{SE: isn't misspelling low-level?} 
and via polarization, i.e., inverting the meaning of the sentences by inserting the word ``not''. %and disclosed weaknesses of it.
\citet{Alzantot2018Generating} introduce an optimization-based algorithm to generate adversarial examples by replacing words in the input. 
%with semantically related words. %and evaluated it on the both \acrshort{nlp} tasks sentiment analysis and textual entailment. 
Their generated words are semantically similar because they are %replaced by %their 
nearest neighbors in the GloVe embedding space.
%\todo{SE: is my change correct?} 
They are also syntactically correct because they need to fit into the surrounding context with respect to the 1 billion words language model.
\citet{Iyyer2018Adversarial} %introduce 
%the \acrshort{scpn} to generate adversarial examples for sentiment analysis and textual entailment. \acrshort{scpn} therefore 
generate syntactically correct paraphrases for a sentence. %obeying a user specified target form. %given a sentence and a target syntactic form. 
%They evaluated their adversaries on two different sentiment analysis datasets and showed that these properly fool the model.
\citet{Ribeiro2018Semantically} %propose semantically equivalent adversaries, i.e. semantically equivalent paraphrases, by translating and back-translating each respective sentence into another language. 
use MT backtranslation to produce %semantically equivalent 
meaning-preserving 
adversaries. They generate adversarial examples for machine comprehension, sentiment analysis and visual question answering to show robustness issues in %three 
\acrlong{sota} models for each task.
\citet{Jia2017Adversarial} insert semantically correct but irrelevant paragraphs into texts to fool neural reading comprehension models.
%\citet{Zhao2017} train a \acrshort{gan} to generate semantically and syntactically coherent sentence adversaries, evaluated different models against it on textual entailment and MT and showed the potential of their adversaries to harm these models.

%\begin{itemize}
    %\item In contrast to adversarial attacks that need further knowledge of s 
    %\item \textit{Adversarial attacks} are the process of inserting modified inputs into a deep-neural network in order to fool the system into a wrong decision \citep{zhang2019adversarial}. 
    %\item At the same time, the original meaning should still be understandable by humans
    %\item The attacks can be separated into high- and low-level
    %\item \citet{eger2019text} has shown the harmfulness of visual perturbed inputs to deep-learning models
    %\item \citet{Belinkov2017synthetic} has letter swapping, keyboard-typos and natural noise extracted from the wikipedia edit history.
    %\item 
%\end{itemize}

\paragraph{Robustness.}
% Adversarial Training
\textbf{Adversarial training} is a commonly used technique to address adversarial attacks \citep{Szegedy2014Intriguing}. %\citet{Szegedy2014} have shown that inserting adversarial examples into the training data makes them somewhat robust to them. 
%\citet{Goodfellow2014} have followed up this approach and introduced the term adversarial training. 
%\citet{Szegedy2014, Goodfellow2014, Ebrahimi2018, Ebrahimi2018a} use the best white-box adversary, i.e., adversarial examples are calculated in respect to the models' parameters, for each time-step in the training.
%In contrast, adversarial training based on black-box adversaries trains on model unspecific adversaries \citep{Eger2019, Alzantot2019, Belinkov2017}.
%Different manners of adversarial training can be used. 
%In the computer vision community, 
%In a white-box setting, 
The term may %often refers 
refer to calculating model gradients with respect to the input and inserting new training examples based on this gradient \citep{Goodfellow2014Explaining}. 
%In black-box settings, 
Alternatively, adversaries obtained from the attacker are inserted at train time \citep{Belinkov2017synthetic,Alzantot2018Generating,eger2019text}. 
\iffalse
How adversarial training is then performed may also vary. E.g., 
\citet{Belinkov2017synthetic} train models on datasets %which are 
perturbed by one attacker and find good performances against the same kind of attack at test time. %in the respective attack scenario. 
They also train on datasets %which are 
perturbed by multiple attackers and find performance increases against all attackers included into the training data. Their best performing model over all attackers on average is the model %which was 
trained on a mixture of all attackers.
\fi % fine

\section{Catalogue of Attacks}
\label{sec:attacks}
%To investigate the effects of different low-level adversarial attacks 
We propose a catalogue of ten different attacks. Our intention is to suggest a maximally inclusive list of potential attacks under the constraint that humans are robust to them.  

\subsection{Attack protocol} Our attack protocol 
%on %RoBERTa 
%deep learning systems 
is \emph{black-box} and \emph{non-targeted} \cite{Xu2019AdversarialAA}: we do not assume access to model parameters and our goal is to fool the system without any desired outcome in mind---in contrast, a spammer would want spam emails to be misclassified as non-spam, but not necessarily the reverse. %situation. 

%We parametrize the attack level by a parameter $p\in[0,1]$. 
%We attack with three different attack levels %with perturbation probability $p$: 
%to model attack scenarios of different strengths. The strength may also affect how well humans %text processing. 
%can recover from an attack. 
We parameterize attack levels by a \emph{perturbation probability} $p\in[0,1]$. 
%and consider the cases of  
%\textit{low} ($p=0.2$), \textit{mid} ($p=0.5$) and \textit{high} ($p=0.8$) attack levels. %to evaluate the model against different strengths of perturbations. Their main difference is the probability to apply a perturbation to a data sample.
With %perturbation probability 
$p$, our goal is to attack $p\cdot 100$\% of all tokens in each sample %, i.e., sequences of tokens $x=(w_1,\ldots,w_n)$, 
in our dataset. %To do so, we go over each sequence $x$ from left to right, token-by-token, and independently flip a coin with tail probability $p$ to determine whether the current token should be attacked. %We do so until 
%\todo{SE: is this approximately what you did?}
To do so, for each sample $w=(x_1,\ldots,x_n)$, %in our datasets, 
we randomly and without replacement draw a token index $i$ to perturb. We independently flip a coin with tail probability $p$ to determine whether the token $x_i$ should be attacked. We do so until either $p\cdot 100$\% of all tokens in $w$ are perturbed or else if there are no more indices left. 

\subsection{Attacks}

Some of our attacks, each of which operates on the character-level of an attacked word, are parametrized by a character-level perturbation probability $\phi$. For simplicity, we set $\phi=p$ throughout, where $p$ is the above defined word level perturbation probability. 

\iffalse
\\
\textbf{Inner-Shuffle} - rotate the positions of the inner characters in a word. 
\\
\textbf{Full-Shuffle} - rotate the positions of every character in a word. 
\\
\textbf{Intrude} - insert random symbols. 
\\
\textbf{Disemvowel} 
\\
\textbf{Truncate} 
\\
\textbf{Segment} 
\\
\textbf{Keyboard-Typo} 
\\
\textbf{Natural Noise} 
\\
\textbf{Phonetic} 
\\
\fi

\paragraph{Inner Shuffle.}
%The first kind of perturbation is the randomization of all letters in a word except the first and last. 
This randomly shuffles all letters in a word except for the first and last.
This attacks %utilizes 
builds on 
the human ability %of humans 
to still comprehend words if the first and last letter remain %untouched 
intact
\citep{Rayner2006RaedingCost}. 
%Due to the omitting of the first and last letter in the randomization process, %change would only occur 
We only allow change in words with length $\geq 3$. %Therefore they are skipped.

\paragraph{Full Shuffle.} %\todo{SE: this attack should not be in the list. No human robustness}
This is the extreme case of the inner-shuffle perturbation %. Here the positions of all letters of a word are randomly exchanged.
where the constraint relating to initial and final letters is dropped. We include this attack for completeness, even though we do not assume high degrees of human robustness to it. 
%We assume that the restoration of short words should not be difficult for experienced readers and because the letters of words of length 1 cannot be exchanged. 
%This attack 
We apply this to all words with length $\geq2$.

\paragraph{Intruders.}
%The third attacker 
Inserting unobtrusive symbols \citep{Hosseini2017Deceiving} in words is a typical phenomenon in social media, e.g., to avoid censorship. Depending on the %placeholder 
symbols chosen, an attack may have little effect on humans.  
%This attack inserts different kinds of punctuation into a word. 
%The inserted symbol is chosen 
We choose the inserted symbol randomly but in case of multiple insertions into one word %it remains the same. %Possible symbols 
keep the symbol identical. 
%are: 
We allow the following symbols to be inserted: 
\( ! " \# \$ \% \& ' ( ) * + , - . / : ; < = > ? @ [ \backslash ] \hat \_ ` \{ | \} ~ \), including whitespace. The perturbation probability $\phi$ additionally influences the number of insertions taking place. For each two characters, $\phi$ indicates how likely the insertion of a symbol between them is. %\todo{SE: is this the same $p$ as below?} 
%Because no symbol can be inserted between words with length 1, this attack only applies to words with length $\geq 3$.
We apply this attack to all words with length $\geq 3$. 

\paragraph{Disemvoweling.}
This %data perturbation 
%will 
removes all vowels (a, e, i, o, u) from a word. If a word only consists of vowels, it will be ignored to prevent it from being deleted. Words with length $\leq3$ are skipped to maintain  readability. %\todo{SE: need a reference}
Disemvoweling is a common feature of SMS language and on social media presumed to require little cognitive effort for humans \citep{Boyd:2010}. %\footnote{\url{https://en.wikipedia.org/wiki/Disemvoweling}}

\paragraph{Truncating.}
This removes a fixed number of letters from the back of a word. 
%to a minimum word length of 3. 
We only cut the last letter from 
%It does not apply to 
words of length $\geq3$ to maintain readability. %\todo{SE: how many are removed?}
Predicting word endings from beginnings is considered an easy task for humans \citep{elman1995language}.
%\todo{SE: truncation needs a psychological motivation - are humans robust to it?}

\paragraph{Segmentation.}
This joins multiple words %are joined 
together into one word. 
%In contrast to the other attacks this perturbation is applied differently. 
Here, the perturbation level is the probability to merge the first two adjacent words. Each following word gets a lower probability to get merged ($\phi^2,...,\phi^n$) to prevent `giant' words.
%The probability to join the $n$th word $w_n$ equates to $p(w_n) = p^n$ where $p$ is the perturbation probability. 
%This attack cannot 
We do not apply this attack to sequence tagging tasks such as \acrshort{pos}, %tagging task, 
because the joined words would have no proper tag, making evaluation more difficult. The ability of humans to segment unsegmented input is already acquired during infancy \cite{goldwater2009bayesian}. 

\paragraph{Keyboard Typos.} %This attack has been taken 
We adopt this attack from \citet{Belinkov2017synthetic} and %has been 
adapt it to our workflow. Hereby, adjacent letters on the English keyboard are replaced by each other randomly. This simulates human typing errors. The higher the perturbation probability $\phi$, the more characters are exchanged by adjacent letters.

\paragraph{Natural Typos.} %This attack has also been taken from \citet{Belinkov2017}. 
Words are replaced by natural human errors from the Wikipedia edit history \cite{Belinkov2017synthetic} %This dataset\footfullcite{charNMT-noise} 
which contains multiple sources of error: phonetic errors, omissions, morphological errors, key-swap errors and combinations of them. %\todo{SE: are these keyboard errors or just a mix of heterogenous typos?}

%%% =======================================================

\paragraph{Phonetic.}\label{par:phonetic}
 An ideal phonetic attack leaves the pronunication of a word intact but alters its spelling. Phonetic attacks are common especially in English with its irregular mapping of pronunciation and spelling. They do not only occur as mistakes but also as a form of creative language use \cite{tagg2011wot}. %\todo[color=green]{YB: In addition we let the original and the phonetic replacement flow through WordSim and only replace it if the model taggs them as \textit{identical} or \textit{very similar}.}

\paragraph{Visual.}
Visual attacks are based on the idea that humans may easily recognize similar looking symbols \citep{eger2019text}. %, as a feature of human perceptual abilities. 
%The visual attacker is based on \textit{VIPER} \citep{Eger2019} and operates on the full input sentence $x$. It takes the probability $p$ and an embedding space, i.e. image-based character embedding space (ICES) computed in section~\ref{sec:visual}, as arguments. 
We replace each character in the input sequence %a perturbation decision is made with respect to $p$. If 
%a replacement should take place the letter is 
%we replaced it by 
with one of its 20 visual nearest neighbors %with respect to the ICES.
%as defined in the so-called `ICES' space in \citet{eger2019text}. 
in the visual space defined below. This attack is also parameterized by $\phi$: we replace each letter in a word i.i.d.\ randomly with probability $\phi$. 

%\todo[inline]{SE: it's not clear to me whether the $p$ used here is the same as the $p$ below. $p$ below applies to words in sequences, but the $p$ in the above attacks applies to something else -- e.g., within the word. Or did you just equate those two $p$?}

\iffalse
\begin{table}
    \centering
    \begin{tabularx}{\linewidth}{lX}
    \toprule
    % begin table header
    \textbf{Attacker} & \textbf{Sentence} \\
    \midrule
    % end table header
    % begin table body
        inner-shuffle & Aadrreavsil aacttks. \\ \hline
        full-shuffle & idaAasvrler tstkaac. \\ \hline
        intrude & A d v e r sar ial at:ta:ck:s. \\ \hline
        disemvowel & dvrsrl ttcks. \\ \hline
        truncate & Adversaria attack. \\ \hline
        segment & Adversarialattacks. \\ \hline
        typo & Adverssrial attaxks. \\ \hline
        natural noise & Adversarial attacs. \\ \hline
        phonetic & Adveraciory attack. \\ \hline
        visual & \customfont{Ǟḋ\textupsilon{}ɝrsarīaɫ aṯtaᴄḳṣ.} \\
    % end table body
    \bottomrule
    \end{tabularx}
    \caption{Ten different modifications of the phrase ``Adversarial attacks.''}
    \label{tab:my_label}
\end{table}
\fi

%\todo{YB: hier ein Absatz?}
We observe that our attacks are \emph{not} directly comparable. For example, at some perturbation level $p$, truncate removes $O(p\cdot n)$ characters, where $n$ is sentence length. In contrast, intruders inserts $O(p^2\cdot n\cdot m)$ characters, where $m$ is a bound on word length. 

\subsection{Implementation of Visual and Phonetic Attacks}

We describe details of phonetic and visual attacks below, as they are more involved.  

\paragraph{Phonetic Embeddings and Attacks.}
%\todo[inline]{SE: can we clarify exactly? Did you not also use Letter2Phon?}
In order to replace words by phonetically similar ones, we use two stages. First, we train two Seq2Seq models to translate a letter string into its phonetic representation and vice versa. We use the Combilex dataset to do so \citep{Richmond2010}. In addition to that, we induce \emph{phonetic word representations}, i.e., a vector space where two words are close if they are pronounced alike. We use an InferSent-like architecture to do so \citep{Conneau2018}. Details are given in the appendix. When a word $x$ should be phonetically perturbed, we run the first Seq2Seq model to obtain a phonemic representation and then convert this back to a letter string $\tilde{x}$ (as in backtranslation in MT). We finally keep %the resulting string when %the obtained word 
$\tilde{x}$ when 
it is phonetically similar to $x$. %the original word. 
We added the latter step because we observed that some resulting words $\tilde{x}$ %were not phonetically similar to $x$ 
had very different pronunciation than $x$ 
after the backtranslation. 

%phonetic word embeddings. %We therefore 

\paragraph{Visual embeddings.} %\subsection{Visual}\label{sec:visual}
% Wie habe ich diese erstellt?
In order to generate %character embeddings 
visual character embeddings, 
we used an architecture introduced by  \citet{Larsen2016} as a combination of \acrshort{gan} and \acrshort{vae}, called \acrshort{vaegan}.  %\acrshort{vae}. 
%A schematic overview of a \acrshort{vaegan} is given in Figure~\ref{fig:vaegan}. 
The model is able to learn embeddings which encode high-level abstract features. %, i.e. wearing glasses, 
%by optimizing a not pixel-based loss function.\todo{SE: what does wearing glasses mean?}
This property is desirable in our case, because humans rely on abstract features \citep{Dehaene2011TheReading}, i.e., shape and spatial relation of the letter, instead of pixels while reading. %The following will give a short introduction to the main aspects of the architecture before we move on to our setup to compute the visual embeddings.
The model is described in the appendix. %\todo{\sout{is it described in the appendix?} ist im Appendix beschrieben}

%Finally, 
To obtain visual character embeddings, we generate a grayscale image of size $24\times{}24$ for each character in the Basic Multilingual Plane (BMP; 65k characters) of the standard Unicode character set with \textit{Pillow}. %\footfullcite{pillow}.
%\todo{SE: some things need clarification here for me} 
%Thereby, we got 65k different images of characters later used as input $x$ for the \acrshort{vaegan}. 
%In our case, the \textit{encoder} applies multiple convolutional and max-pooling steps to the image to compress it. 
%The \textit{decoder} and \textit{generator} apply multiple deconvolutional and sampling steps respectively. 
The \acrshort{vaegan} is trained on the full BMP dataset. 
After training, we compute 256-dimensional visual letter embeddings by encoding the respective letter image with the encoder of the \acrshort{vaegan}.
The quality of the embeddings can be derived via the models’ ability to properly reconstruct an image from them, see  Figure~\ref{fig:vaegan_reconstruction} in the appendix. 
%The models' capability to properly reconstruct an image from its embedding is shown in Figure~\ref{fig:vaegan_arch} in the appendix. \todo{YB: Better formulation? Because the embedding should be in the focus not the model itself. The quality of the embeddings can be derived via the models’ capability to properly reconstruct an image from its embedding shown in Figure~\ref{fig:vaegan_arch} in the appendix}%The better the model has encoded the visual features of the image the more identical the two pictures are.

%\todo[inline]{SE: can you please check the error message for this figure? Also, one figure is missing and both of them should go to the appendix. YB: Woher der Fehler kommt weiß ich noch nicht so genau aber die Figure Referenzen sollten korrekt sein. Es wird nur irgendwas abgeschnitten deshalb kommt es wohl zu diesem Verhalten.}

%[todo] % obv fine

\section{Experimental Setup}
\label{sec:experimental_setup}
\subsection{Base model and datasets}

\begin{table*}[!htb]
{ %\small
    \centering
    \addtolength{\leftskip} {-2cm}
    \addtolength{\rightskip}{-2cm}
    \begin{tabular}{llcccc}\toprule
        \thead{Task} & \thead{Dataset} & \thead{Train} & \thead{Test} & \thead{Clean score} \\%& \thead{Best score from literature} 
        \\ \midrule
        \thead{\acrshort{pos} Tagging} & Universal Dependencies (part) & 13k & 2k & 96.95 \\ %& 95.59\% \\ %\midrule
        %\thead{\acrlong{nli}} 
        \thead{NLI}
        & \acrlong{snli} & 550k & 10k & 90.41 \\ %& 91.9\% \\ %\midrule
        \thead{Multilabel} & Toxic Comment & 560k & 234k & 0.93 \\ %& 0.98 \\\\
        \thead{Classification} & \\
        \bottomrule
    \end{tabular}
    \caption{Overview of the \acrshort{nlp} tasks used in this work. 
    Clean scores are scores from training and testing on clean data. 
    %Additionally, the data splits are given. %and the \acrshort{sota} performance are given. These are measured in accuracy for \acrshort{pos} and \acrshort{nli}, and in AUCROC score for multilabel-classification.}
    }
    \label{tab:datasets}
}
\end{table*}

Our base architecture used in all experiments is %the \acrshort{sota} transformer model 
\textit{RoBERTa} \citep{liu2019roberta}. %It is pretrained on a Masked Language Modeling task whereat tokens in the input get masked and it has to recover the true input.
%Thereby it learns contextualized word embeddings which can later be used by downstream tasks (e.g. sentiment analysis). 
%High performances has been shown on a variety of downstream benchmarks: GLUE \citep{Wang2019a}, RACE \citep{Lai2017} and SQuAD \citep{Rajpurkar2016,Rajpurkar2018}. 
RoBERTa is a robustly optimized extension of BERT that has been trained (i) for longer, (ii) on more data, and (iii) without the next sentence prediction task. RoBERTa has been shown to outperform BERT on a variety of benchmark tasks, including those contained in GLUE \citep{wang-etal-2018-glue}. 
We study the performance of \textit{RoBERTa} in our attack scenarios %we evaluate it 
on three different \acrshort{nlp} tasks. %with their respective 
Dataset statistics are shown in Table~\ref{tab:datasets}.

\paragraph{\acrshort{pos} tagging} is a %word level 
sequence tagging task where %the model has to label 
each token in the input needs to be labeled with its respective \acrshort{pos} tag. %(e.g. \textit{ADJ}, \textit{NOUN}, \textit{VERB}). 
We use the English universal dependency dataset with 17 different tags %as our dataset 
\cite{Nivre2016Universal}. %\todo{\sout{SE: reference missing} fixed}  
%The universal dependencies dataset used in this task consists of 17 different tags \cite{}. %The best performance reached on the full English portion of the dataset is 95.59\% accuracy score \citep{Cui2019}.

\paragraph{\acrshort{nli}} is a %sentence level 
classification task in which %the model has to predict 
the relation of a sentence pair %of sentences 
must be predicted. %Possible labels are: 
Relation labels are 
\textit{neutral}, \textit{contradiction} and \textit{entailment}.        %The 
We use \acrshort{snli} \cite{Bowman2015}. %as dataset. %dataset is used for this task. %The current \acrshort{sota} performance is by \citet{Zhang2019} with 91.9\% accuracy score\footfullcite{snli_website}.
%\todo{SE: are you not using a subset??\\YB: No}

\paragraph{Toxic Comment Classification (\acrshort{tc})} 
%is a sentence classification task which %is a natural candidate %for attack scenarios in NLP due to its role in social media. 
%for attack scenarios. 
%may naturally include textual attacks as users in social media platforms who want to post offensive content may want to prevent being detected and censored.  
labels sentences (typically from social media platforms)
%The task is to label a sentence 
%with one or multiple labels. %\todo{SE: I think one can label each sentence with multiple labels, not only one} 
%into possible 
with one or several toxicity classes.
Possible labels are: \textit{toxic}, \textit{obscene}, \textit{threat}, \textit{insult} and \textit{identity hate}. For this task, we choose the jigsaw toxic comment challenge dataset from kaggle\footnote{https://www.kaggle.com/c/jigsaw-toxic-comment-classification-challenge}. %\todo{YB: check if reference is ok}}%\footfullcite{jigsaw_toxic_comment_classification}
The current best performance on the leaderboard has an AUCROC (area under the receiver operations characteristic curve) score of 98.8\%. %\footfullcite{tc_leaderboard}
%AUCROC measures how much the model is capable of distinguishing between the different classes. %\todo[inline]{SE: AUCROC is area under the curve. Maybe it should be defined differently}
%In addition this task is a good application scenario because it has practical relevance (e.g. spam filters, hate speech detection).

%----------------------------------

\iffalse
In order to perturb the datasets, we apply the following process to each data sample.
Let $x=(w_1,\ldots,w_n)$ be the token sequence of a sample %can be represented as $x = [w_1,...,w_n]$. 
in a dataset. 
%We calculate the number of tokens to be perturbed $t_{target}$ by multiplying the token count $|x|$ with the perturbation probability $p$.
We multiply $n$ with the perturbation probability $p$ to determine how many tokens in $x$ should be attacked. 
Afterwards, we pick a random token $w_i$ and perturb it. 
%if possible (e.g. if a phonetic perturbation of the token is existent). 
We repeat this process until we perturbed $t_{target}$ tokens or every token in the input has been considered. 
\fi
%\todo{SE: where does $p$ come into play?} 

%We perturb each downstream task dataset introduced in section~\ref{sec:task_data} with the three different attack levels and one perturber. 
%We repeat this for every perturbation listed in the following and therefore get 30 different datasets for each task.

%\section{Robustness of RoBERTa}
\subsection{Results}
%In order to figure out which impact out attacks have on existing SOTA DL models we perturbed test data for three different tasks: POS tagging, SNLI, and toxic comment classification.
%As can be seen in \ref{fig:adversarial_performance} the performance decreases under attack.

\begin{figure*}[!htb]
    \centering
    \includegraphics[page=1,width=\textwidth]{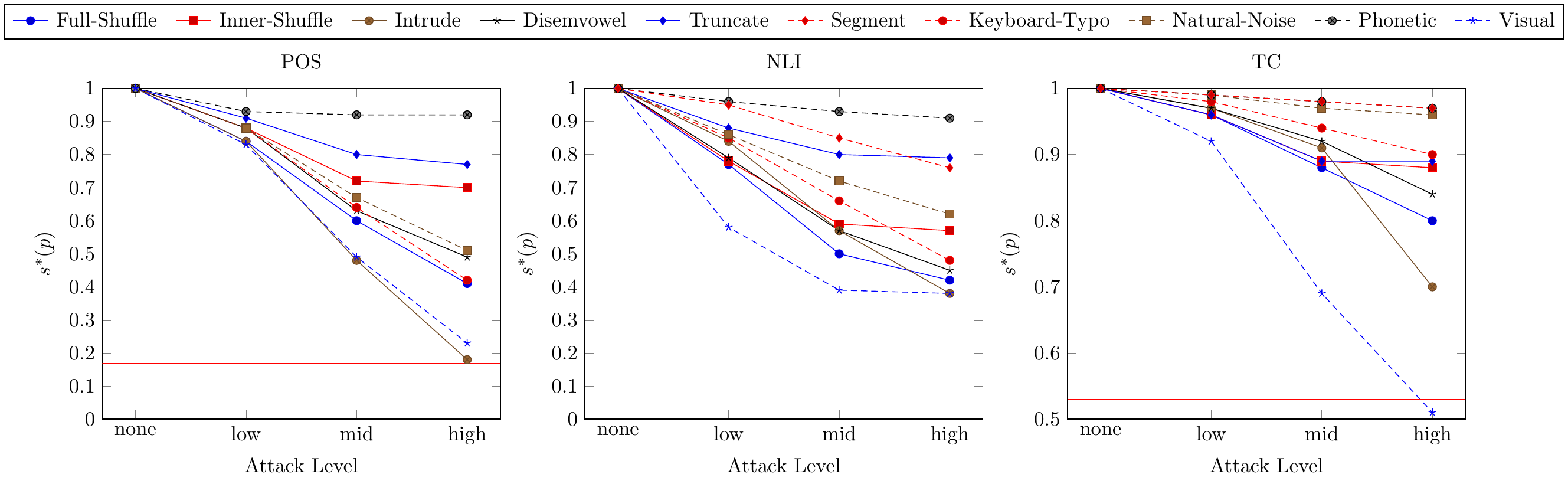}
    \caption{Performance decreases of RoBERTa on the three downstream tasks: POS, SNLI an TC. Red lines indicate the random guessing baseline. %performance (e.g. random guessing of classes) that can be reached}
    }
    \label{fig:adversarial_performance}
\end{figure*}

%\begin{figure*}
%    \centering
%    \includegraphics[page=3,width=\textwidth]{figures/figures.pdf}
%    \caption{Leave-One-Out robustness increase compared to clean model.}
%    \label{fig:leave-one-out_robustness}
%\end{figure*}

%\todo{SE: why did you include identical figures twice in your thesis?}
We consider the cases of  
\textit{low} ($p=0.2$), \textit{mid} ($p=0.5$) and \textit{high} ($p=0.8$) attack levels.

In Figure~\ref{fig:adversarial_performance}, we plot the performance of \textit{RoBERTa} for the three tasks \acrshort{pos}, \acrshort{nli} and \acrshort{tc} individually as we perturb the test data using our attackers. %For detailed data consult 
Detailed numbers are reported in 
Table~\ref{tab:attack-performance} in the appendix. %The x-axis represents the different attack levels, the y-axis reports the relative performance equal %as in \citet{eger2019text}
We report scores relative to the model performances on the clean test set: 
\begin{equation}\label{eq:s_p}
    s^{*}(p)=\frac{s(p)}{s(0)}, \quad p\in\{0,0.2,0.5,0.8\}
\end{equation}
where $s(0)$ is the task specific performance on clean data listed in Table~\ref{tab:datasets} %\ref{tab:clean_performance} 
and $s(p)$ is the performance %under attack with 
for attack level $p$. %$s^{*}(p)$ is 
Scores are measured in accuracy for \acrshort{pos} and \acrshort{nli}, and in AUCROC for \acrshort{tc} classification. % An important note about the performance measurements is that they depend on their respective task and dataset. 
Clean performance scores depend on the specific task and dataset. 
%That means the 
For example, \acrshort{nli} %reaches its worst performance at around a score of 
has a worst score of around $33\%$ accuracy (majority label) %, because it assigns the majority class label to each input, 
and \acrshort{pos} has a corresponding worst score of around $16\%$ accuracy. %Additionally 
The worst performance of \acrshort{tc} is reached at AUCROC score of $50\%$---at this point, the model is no longer able to distinguish between the different classes. We mark these values relative to the tasks' best performance ($s(0)$) in Figure~\ref{fig:adversarial_performance} as red lines. %\todo{SE: not immediately clear to me why you divide again by $s(0)$} 
Each task suffers performance decreases from each attacker. 
The higher the perturbation level, the lower the model performance. %In the following, the respective task-specific model is referred to as `the model'.

% Phonetic
The \textbf{phonetic} attack is the least effective for all tasks with maximally 10 percentage points (pp) %\todo{change to: pp?} 
performance decrease %as its best 
with the \textit{high}est perturbation probability of $0.8$. % We assume the low performance is because of the small changes applied due to the few possibilities to properly perturb a word phonetically.
%Additionally, 
%but the effect of the three different attack levels almost does not differ. % because some words do not have a phonetic perturbation. Therefore they cannot be replaced and the count of perturbed samples per sentence remains mostly the same while increasing the attack level.
% Truncate
The \textbf{truncate} attack yields %better results 
higher performances decreases 
%than the phonetic attack 
in all three tasks, %It is %therefore somewhat 
being 
roughly twice as effective. %We assume this is a result of the hyperparameter selection. The small number of characters that were truncated seems to be too small to properly fool the model into a wrong decision. 
The performance decreases by 10pp from \textit{none} to \textit{low} and additional 10pp from \textit{low} to \textit{mid}. %\todo{SE: 10 points or 10\%?} %Furthermore the increasing attack level from \textit{mid} to \textit{high} has almost no impact on the performance on the downstream task. 
Increasing the attack level beyond that does not cause further harm, especially for \acrshort{nli} and \acrshort{tc}. 
%Especially the sentence level tasks do not suffer noticeably more by increasing the perturbation level to \textit{high}. % We assume this is due to \acrshort{nli} and \acrshort{tc} are operating on a more abstract sentence understanding, e.g. the occurrence of keywords to accomplish their task, in contrast to \acrshort{pos} tagging which, as a word-based task, relies a lot on each word and its surrounding context.
% Segmentation
%In addition the sentence based tasks were evaluated against the \textbf{segmentation} attack. 
Concerning the \textbf{segmentation} attack, 
for \acrshort{nli}, it leads to a similar performance decrease 
%for the model 
as the truncate attack for \emph{small} $p$, %In contrast it gets linearly 
but becomes more successful as the perturbation level increases 
%from \textit{low} 
to \textit{mid} and \textit{high}. For \acrshort{tc}, the performance decrease is almost identical to the phonetic attack. %The low performance decreases in both tasks seems to result from \textit{RoBERTa}s' BPE\footnote{\textbf{B}yte-\textbf{P}air \textbf{E}ncoding relies on subword units \citep{Liu2019}} tokenizer. The model is therefore able to split the segmented words back into its parts, i.e. the original words and their subwords, to gain a rough understanding of the original sentence. Thereby the original word meanings should still be understandable by the model. 
%\acrshort{nli} suffers a lot more than \acrshort{tc} by around 25\% rather than 5\%. % - we assume that hidden features that are left intact by the attack, e.g. sequence length or occurrences of keywords, on which the model bases its decision for the respective task make the difference.

% Natural
We notice a linear decrease in performance for each task when increasing the perturbation level of the \textbf{natural-noise} attack. %In \acrshort{tc}, it has a similar low impact to the model as the segmenation attack by around 5\% for \textit{high}. This is in contrast to 
Especially \acrshort{pos} and \acrshort{nli} %which 
suffer a %decent 
strong performance deterioration of around 40pp and 50pp for the \textit{high}est attack level. Both lose 15pp to 20pp performance per attack level increase. %We assume the stark contrast depends on natural occurrences of this kind of perturbation in \acrshort{tc}s' original train dataset, because it is crawled from real world examples, e.g. social media comments. Therefore it seems like the \acrshort{tc} dataset somehow applies unintended adversarial training to the model and makes it perform clearly better in this attack scenario compared to \acrshort{nli} and \acrshort{pos}.

% Full & Inner Shuffle
\textbf{Full-} and \textbf{inner-shuffle} %are applying roughly the same kind of perturbation to the data. 
randomize the order in an input word but humans are more robust to inner-shuffle. Full-shuffle also affects \textit{RoBERTa} more than inner-shuffle. It tends to be one of the strongest attack scenarios, while inner-shuffle typically ranks in the midfield.  
%In general all tasks are suffering from both methods. 
\iffalse
In the \acrshort{pos} and \acrshort{tc} task the performance of the model decreases linearly as the full-shuffle attack level increases to a maximum at probability \textit{high} where it degrades by 40\% and 20\% relative performance respectively. For \acrshort{nli}, the performance deficit is around 25\% each for the increment from \textit{none} to \textit{low} and \textit{low} to \textit{mid} and lowers to 10\% for the transition to \textit{high} to a maximum of 60\%. In contrast the inner-shuffle reduces the models' performance linearly while increasing the level from \textit{none} to \textit{low} and \textit{mid} for all tasks. In detail by 15\% for \acrshort{pos}, 20\% for \acrshort{nli} and 5\% for \acrshort{tc}. Increasing the level to \textit{high} seems to have a low impact.
\fi
% We assume the probability increment does not affect the success of the inner-shuffle attack beyond the \textit{mid} attack level and its maximum effect has been achieved. The major performance difference between the two types of perturbation seems to be related to the first and last letter and therefore to the count of letters which are shuffled. A possible explanation is that the first and last letter have a special influence for the models' understanding of the respective word and therefore on the overall performance of the task.

% Disemvowel
The \textbf{disemvowel} attack has different effects %on model performances 
in different tasks. 
For \acrshort{pos}, it is almost identical %nearly aligns 
to the natural-noise attack with a slightly stronger  %performance 
impact of 5pp for \textit{mid} and 3pp for \textit{high} and a maximum on 50pp. 
\acrshort{nli} loses around 20pp performance on \textit{low} and it decreases an additional 20pp by increasing the level to \textit{mid}, %while reaching 
and reaches 
its greatest decrease by 55pp on \textit{high}. In \acrshort{tc}, %the 
model performances decrease linearly from \textit{none} to \textit{low} and \textit{mid} by 8pp each. The \textit{high} attack level doubles %this decrease 
to a total of 15pp performance loss. 
% Intrude
The \textbf{keyboard-typo} attacks have median impact throughout tasks and attack levels. %, but generally is more impactful than disemvowel. 

The \textbf{intrude} attack %is one of the best performing attacks for all three tasks. 
is among the most severe attacks across all three tasks. 
For \acrshort{tc},  the \textit{low} and \textit{mid} attack levels have a relatively low impact compared to \textit{high} which yields a performance loss of 30pp. % We assume this is due the \acrshort{tc} dataset often consist of long texts with several sentences. After perturbing the data, the words are unknown to the tokenizer and are therefore encoded based on their characters. This means that the number of tokens increases significantly. Thereby \textit{RoBERTa}s' maximum sequence length is exceeded and the end of the input is cut off. This may cause the model to lose important information. This effect also scales with the attack level and would therefore explain the strong drop in performance with the \textit{high}est level. In the \acrshort{nli} task the model is losing about 20\% performance each attack level increase to a maximum of 60\% performance loss. This degradation indicates the models' inability to properly perform the task. 
It decreases %the 
model performance the most on the \acrshort{pos} task by above 80pp. 
%It still performs marginally better for the \acrshort{pos} task. % We assume this is because the attacker spared some short words and punctuation for which the model is still able to predict the correct tag.
% key
%linearly decrease %the 
%model performances with respect to the attack level. \acrshort{nli} and \acrshort{pos} start on around 15\% total decrease on \textit{low} and decrease further to 35\% for \textit{mid} and 55\% for \textit{high}. Therefore this attack has a high impact on the model. In contrast the performance decreases for the \acrshort{tc} task are lower compared to the other two tasks. It only drops to a maximum of 10\% at the \textit{high}est attack level. % We assume the model is marginally resistant because human writing errors that are similar to the perturbations generated by this attack naturally occur in the training data.
% Visual 
Especially for both sentence-based tasks \acrshort{nli} and \acrshort{tc}, the \textbf{visual} attack decreases %the model is 
are also among the most severe, while  
%nearly to its worst possible performance. %These results indicate that it has lost its understanding and is therefore randomly guessing classes. 
%The models 
\textit{RoBERTa} is marginally more robust on the \acrshort{pos} task. 
%Important to note is that 
Even for the \textit{low} perturbation level, the \acrshort{nli} model suffers from more than 40pp performance decrease. %It maximizes with perturbation level \textit{mid} on 60\% performance decrease. 
%For \acrshort{tc}, the \textit{low} attack level only impacts the models' performance by 10\%. 
%This increases to 30\% for \textit{mid} and 50\% for \textit{high}. 
The performance for high $p$ even falls below the red line marked as our %worst performance possible. 
lower bound baseline. 
%This is possible because the AUCROC is defined in the interval $[0,1]$. Values below $0.5$ indicate that the model confuses the labels in a way it predicts 0 for 1 and vice versa.\todo{SE: why is this? Maybe because you have a double $p$?}
%\todo{SE: is there any logic in the ordering of the discussion?}

% TODO: Zusammenfassenden Teil über die Attacks schreiben.
%\acrshort{tc} deals a lot better with most of the attacks especially because perturbed data is part of the original dataset and therefore it is trained on such samples. This can be described as a kind of adversarial training against different kinds of perturbations. This does not make it resistant against all types of attacks as can be seen for the visual, intruder, disemvowel and the full-shuffle attacks.
%The \acrshort{pos} model retains a large part of its performance compared to the other two tasks. Overall this seems to be correlated with the fact that \acrshort{pos} tagging is a word-based task. Therefore some words in the perturbed input still remain untouched and keep the performance above a certain threshold.

\subsection{Defenses}\label{sec:res_defenses}
%In the following figures the performances of the three different tasks are reported in the same way as already mentioned above. 
%We replace the value on the y-axis by the performance of our methods to make the model more robust as relative difference equal to \citet{eger2019text} and defined as:
In the following, we report the performance increase from shielding the methods with adversarial training: %\todo{SE: the adversarial training needs to be described} 
\begin{equation}\label{eq:delta}
    \Delta_{\tau}(p):=\frac{\sigma(p)}{s(0)}-s^{*}(p)
\end{equation}
where $\sigma(p)$ is the score for each task with one of two  defense methods $\tau$: %introduced in 
%section~\ref{sec:defenses}.\todo{SE: introduced where?} 
%\todo[inline]{SE: @YB: can you document the adversarial training here in green for me to check}
\begin{itemize}[topsep=0pt,itemsep=-1ex,partopsep=1ex,parsep=1ex]
    \item \textbf{1-1 adversarial training($\alpha$,$\beta$)}: Here, we train on a mixture of %roughly 1/3 
    %data attacked 
    \textit{low}, \textit{mid}, \textit{high} attacked data (each perturbation level is roughly equally likely to appear in the training data). We attack with some attacker $\alpha$ and measure performance when the test data is attacked with attacker $\beta$. 
    \item \textbf{leave-one-out} (LOO): Here, we train on a mix of all attackers except for the one %attacker 
    with which the test data is attacked. The train data contains an equal mix of data from each attacker and attack level.  
\end{itemize}
%\textcolor{ForestGreen}{In 1-1 hab ich auf einem Mix aus den unterschiedlichen Attack Leveln trainiert. Also in etwa 33\% low, mid \& high. Für 1-1 insb. auch nur auf einem Perturber und danach auf allen evaluiert deshalb ist es in den Results ne Matrix. \\
%\\
%Fuer LOO habe ich das gleiche mit den leveln gemacht nur dass ich alle perturber gemischt habe. Also von jedem Perturber (ausser von dem auf dem evaluiert wird) sind ungefaehr 10\% der Gesamtdaten die fuer das AT benutzt wurden.}
%Therefore 
%For example, a score of 0.3 indicate 30\% performance (robustness) improvement compared to the non-shielded setting. %model for a respective task, attacker and level. 
%It is important to mention that an improvement in the following do not take place from one attack level to another but relative compared to the performance of the non-shielded model against the respective attacker. 
%We describe the performance differences of each shielding approach individually. 

\subsubsection{Adversarial Training}
\paragraph{1-1 ($\alpha,\alpha$)}

\begin{figure*}[ht]
    \centering
    \includegraphics[page=2,width=\textwidth]{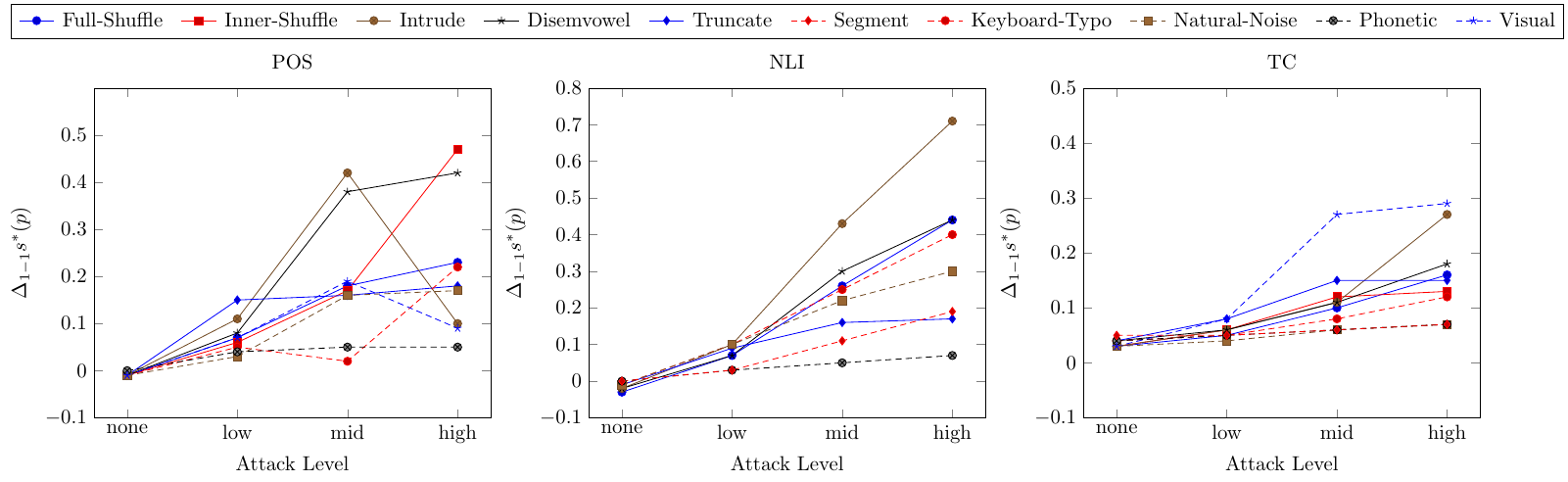}
    \caption{Performance improvements of the models adversarial trained and evaluated individually on the attacker introduced in Section \ref{sec:attacks} for \acrshort{pos} left, \acrshort{nli} mid and \acrshort{tc} right. Performance measured in $\Delta_{\tau}(p)$ defined in Eq.~\ref{eq:delta}.}
    \label{fig:at_1-1}
\end{figure*}

In Figure~\ref{fig:at_1-1}, we report the performance of our models each trained on perturbed data and evaluated against \emph{the same kind} of perturbation. %, which we refer to as 1-1. 
This %generally 
gives 
an unrealistic upper bound %, but is unrealistic, 
since the defender would have to know how it is being attacked.   
%The adversarial trained model on the respective task is mentioned in the following as the model.

% ==================================
%\todo[color=green]{YB: figure 1 \& 3 ist 2x im Paper. SE: fixed.}
% ==================================

% POS
%The first thing to notice 
For \acrshort{pos}, %is that 
the adversarially trained models lose a %little 
bit of their performance on clean data, %and that overall the models' performance against the proposed attackers gets better. 
but their performance on perturbed data improves, especially  
%On \textit{low} the performances 
against intrude and truncate for the \emph{low} attack level. %have to be particularly mentioned. 
%Against the truncate attacker the robustness of the model increases by 15\% and for intrude by 10\%. 
The robustness improvements for the remaining attackers are very similar and range from 3pp increase for the natural-noise attack to 8pp for the disemvowel attack. With one exception, the improvement at large perturbation levels $p$ is highest, and obtains a maximum improvement of 40pp for inner-shuffle. 
For \acrshort{nli}, %we do not report inner-shuffle and visual. 
%This is because we were unable to train the models to perform the \acrshort{nli} task on the perturbed dataset. 
%except for phonetic and segment the models' performances drop by to -3\% on clean data. 
the models %performances 
again tend perform worse on clean data. %\footnote{We do not report on inner-shuffle and visual [BECAUSE?]\todo[color=green]{für nli hab ich alles versucht aber das modell wollte einfach nichts lernen -> random guessing/majority label (bin ich mir aktuell unsicher)}} 
As the perturbation level increases, we see a smooth and steady increase of the values $\Delta_{\tau}(p)$ across all attackers. Improvement is best for intrude which was also among the most damaging attacks.  
%Additionally, all performances increase strictly as the attack level increases. 
%For \textit{low}, the total performances increases stay close together. 
%Against phonetic and segment, the performance increases remains unchanged. 
%In contrast, the remaining models increase their performance to 8\% on average. %, e.g. natural-noise to 10\%. 
%Especially for attack level \textit{mid} the performance against the intrude attack increases significantly to 43\%. 
%The performance against disemvowel increases to a total improvement of 30\% and the performance against the natural-typo to 22\%, respectively. 
%The performances against full-swap and keyboard-typo increase to around 25\%. 
%The models' performance against truncate and segment increases to 18\% in average. 
%The performance against the phonetic attack again increases nearly imperceptible by 2\%.
% zusammenfassende worte für nli
% So in the end we can say that \acrshort{nli} overall performs better against the attackers if trained on perturbed data beforehand. Especially for the strong performing attackers adversarial training seems to be particulary succesful for \acrshort{nli}.

% TC
%As a specialty in the \acrshort{tc} task the models' performances gain a performance increase on clean data of 4\% in average. 
For \acrshort{tc}, model performances increase also on clean data, which is likely due to the nature of the task. 
%For %attack level 
%\textit{low} attack level, only the performance increase of visual and truncate stand out slightly to a total of 8\%. 
%The performances against the remaining attackers increase to 5\% on average. 
%For \textit{mid} the gap between visual and the remaining attackers enlarges significantly. 
%The performance against the visual attack is increasing to 27\%. 
%In contrast, performances against phonetic and natural settle down below. 
As the attack level increases, %there is a gradually increasing robustness 
$\Delta_{\tau}(p)$ gradually further increases across tasks. 
%The performances against the remaining attackers increase to 11\% on average. 
For \emph{high}, largest increase is again observed for intrude as well as for visual, which also had largest impact in the non-shielded setting.  
%For \textit{high} the performance against the visual attacker flattens its increase and remains on 29\% performance improvement in total. 
%A clear performance increase is accomplished against the intrude attack to 28\%. 
%The performances against truncate and inner-shuffle stay nearly unchanged and the remaining performances keep their behavior and increase to 15\% on average.
% Zusammenfassende Worte für tc
% \acrshort{tc} is the only task where the model benefits performance increases even on clean data. These cause that the results in general are better. % We assume that for the \acrshort{tc} task the model also has to cope with perturbed data in general and therefore 

\begin{figure*}[!htb]
    \centering
    \includegraphics[scale=0.6]{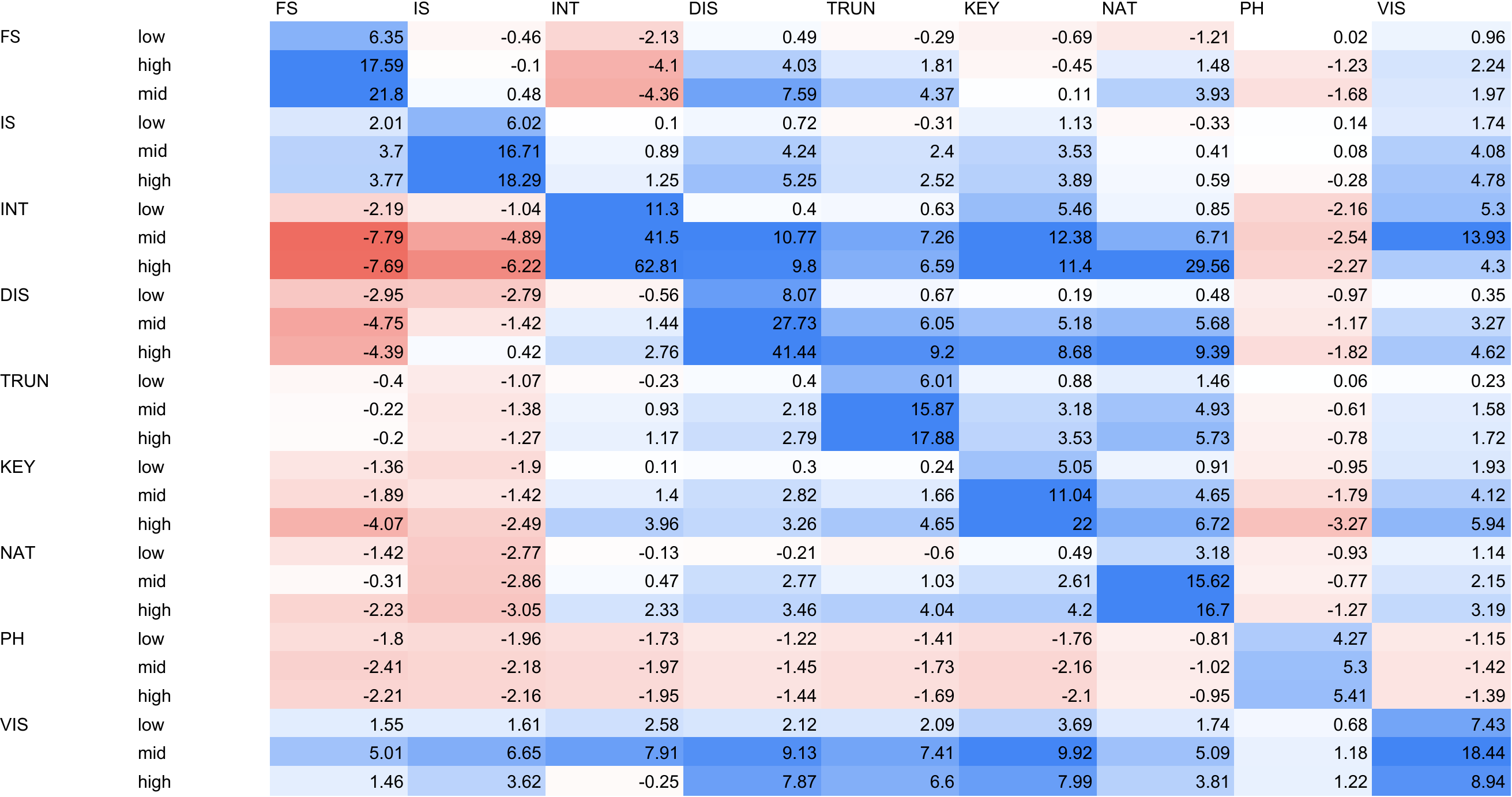}
    \caption{1-1 $(\alpha,\beta)$ adversarial training for \acrshort{pos}. Column: train, row: test. Numbers give values  $\Delta_\tau(p)$, see Eq.~\eqref{eq:delta}. Red colors give performance decreases, relative to the results on clean data; blue colors show increases.}
    \label{fig:adversarial_training}
\end{figure*}

% Zusammenfassende Worte über 1-1
%\todo{In p5:615-618, why does the order of similar adversarial attack groups starts from (2)?; The authors should write "(1) removes or replaces characters" first, to make the coreference coherent.}
\paragraph{1-1 ($\alpha,\beta$)} In Figure \ref{fig:adversarial_training}, we show all 1-1 values for different combination of attackers on train ($\alpha$) and test data ($\beta$). We see that the diagonal ($\alpha=\beta$) always profits considerably, but the off-diagonal can be positive or negative, depending on the choice of $\alpha$ and $\beta$. We clearly see that (1) truncate, disemvowel, keyboard-typo, natural noise,  visual, and intruders are similar in the sense that training on them shields against their attacks at test time. (2) Full-shuffle and inner-shuffle form a second group and (3) phonetic attacks a third group. This is to some degree a natural clustering, as (1) removes or replaces characters, (2) destroys the order of words, and (3) modifies entire words using more complex operations. visual is an outlier in group (1), since it improves no matter what attacks are added at train time.   

\paragraph{Leave-One-Out}

\begin{figure*}[ht]
    \centering
    \includegraphics[page=3,width=\textwidth]{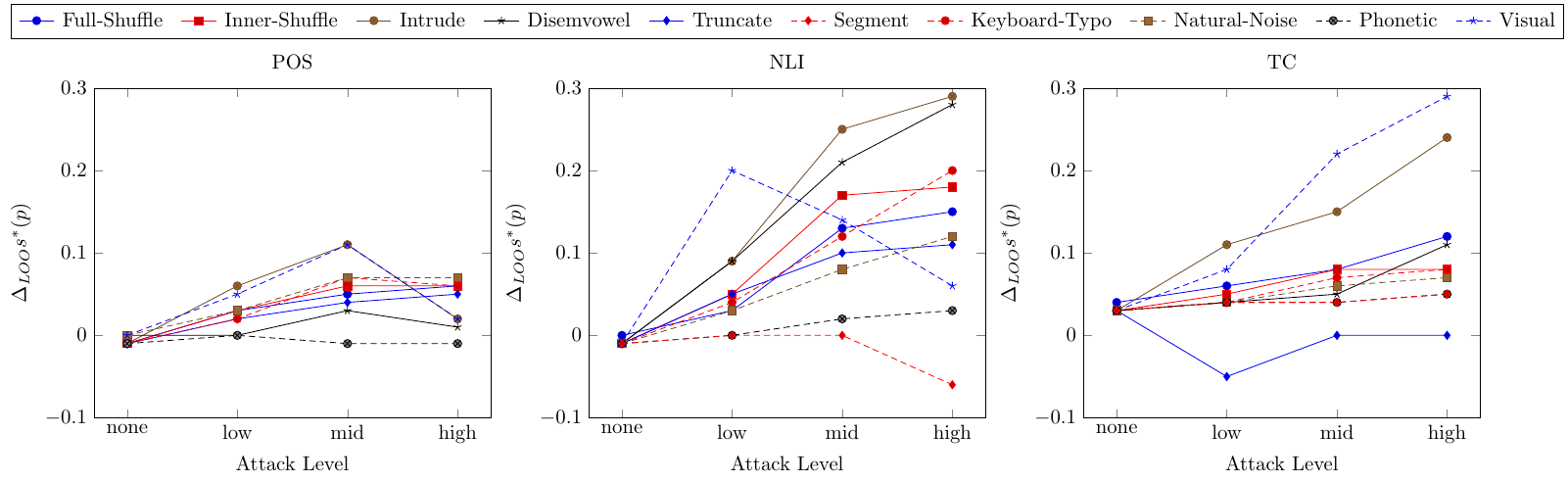}
    \caption{Leave-one-out defense: Performance improvements of the models adversarial trained on all attackers introduced in Section \ref{sec:attacks} except the one they are evaluated on for \acrshort{pos} left, \acrshort{nli} mid and \acrshort{tc} right. Performance measured in $\Delta_{\tau}(p)$ defined in equation~\ref{eq:delta}.}
    \label{fig:at_loo}
\end{figure*}

Figure~\ref{fig:at_loo} %we report 
shows the performance of our models when trained on a mixture of all attackers except the one %it is 
evaluated on. %This is the most realistic scenario of model defense in the case of an unknown new low-level attack scenario at test time.\todo{SE: it needs to be described how defense is implemented in this situation. YB: siehen oben}
This is the most plausible scenario of model defense in the case of an unknown new %low-level 
attack scenario at test time. 

%In figure~\ref{fig:at_loo}
% ==================================
% ==================================

% POS
For \acrshort{pos}, %there can be differentiated between three groups that behave similar. 
the performance against the phonetic attack remains mostly unchanged, while %The models' performance 
$\Delta_{\tau}(p)$ increases as a function of $p$ against natural-noise, inner-shuffle, full-shuffle, truncate and keyboard-typo. 
%improves with increasing attack level. 
The best %performance of this group 
defense is %accomplished 
against natural-noise with 3pp for \textit{low} and 7pp for \textit{mid} and \textit{high}. 
%The performances against the remaining three 
Shielding against %attackers 
visual, intrude and disemvowel attacks %are characterized by a significant 
yields lower values $\Delta_{\tau}(p)$ on attack level \textit{high} compared to %the value on 
\textit{mid}. Overall, we see mild improvements compared to the unshielded situation, but expectedly, these are lower than for 1-1 shielding.   
%The performance against disemvowel differs from the other two by only increasing to 5\% instead of 12\%.

% NLI
For \acrshort{nli}, the performance against keyboard-typo, full-shuffle, inner-shuffle, natural-noise and truncate exhibits steady improvements with increasing attack level which range from 10pp to 20pp for attack level \textit{mid} and \textit{high}. The performances against intrude and disemvowel also show steady improvements with the attack levels %whereas these improvements are significantly 
but are generally higher with up to 29pp. For attack level \textit{low}, the performance improvement against the visual attacker is with 20pp more than twice the value of the others. This improvement diminishes in the \textit{mid} and \textit{high} attack levels and even drops there below the improvements against most of the other attackers. %Particularly noticeable is the deterioration against segment at attack level \textit{high} to -6\%.

% TC
In the \acrshort{tc} task, %the models' performance against the visual and intruder attack attract attention. 
the performance against visual improves even for \textit{low} level to 8pp, increases for \textit{mid} to 23pp and maximizes to 29pp total improvement for attack level \textit{high}. 
The performance against the intrude attack is also very good: for  \textit{low} attack level %\textit{low} 
%indicates 
%yields a higher value (11\%) compared to the visual improvement (8\%). 
the improvement (11pp) is even higher compared to visual (8pp). 
%However, it does not result in a performance increase as high as against visual and only increases to 15\% for attack level \textit{mid} and 25\% for \textit{high} respectively. 
The performances against full-shuffle, inner-shuffle, disemvowel, segment, keyboard-typo, natural-noise and phonetic behave similar for attack level \textit{low} and \textit{mid} with 4pp to 7pp total improvement. 
Shielding against full-swap and disemvowel is slightly better than the last group. %further increase %compared to the before mentioned to around 12\% total increase. 
%The previously inconspicuous performance against the truncate attack deteriorates by 5\% for attack level \textit{low} and remains unchanged for \textit{mid} and \textit{high}.
There is no overall positive effect for truncate. 

 % something is wrong with floats here
\subsection{Discussion}\label{ch:discussion}
%\textcolor{ForestGreen}{
%In chapter~\ref{ch:results} we have shown the impacts of different attackers to the models' performance on the three individual downstream tasks \acrshort{pos}, \acrshort{nli} and \acrshort{tc}. In this chapter we analyse and discuss them.
%}
%\subsubsection{Attacks}
Overall, the phonetic attack was least effective. 
%in our experiments. 
We assume this is because few words were changed overall %[as they do not have a phonetic replacement]\todo[inline]{SE: to understand this, one first needs to define how you attacked phonetically}
as a considerable amount of phonetic replacements were either identical to the input and some were even discarded. 
The truncate attack performed better than the phonetic attack in all three tasks %. It 
%achieved roughly twice the performance reductions. However, we expected results to go beyond that. We suppose this is due to the fact that we chose 
but it still remained low overall, possibly as we truncated only by 1 character, leading to small changes in the appearance of a word.  
%the number of characters that are truncated too small. 
%This leads to too minor changes that apparently do not fool the model into a wrong decision.
%}

% Segmentation
%\textcolor{ForestGreen}{
%The two sentence-level tasks were also evaluated against the segmentation attack. 
We attribute the low impact of the segmentation attack to \textit{RoBERTa}'s BPE encoding, which apparently allows it to partly de-segment unsegmented input. We observe that some attacks (e.g., segmentation, keyboard-typo, and natural-noise) have less effect in \acrshort{tc} compared to \acrshort{pos} and \acrshort{nli}, possibly %We attribute this to 
because of higher natural occurrences of these phenomena in the \acrshort{tc} dataset. %compared to the other two. 
\checkme{The intrude and visual attacks are among the strongest. This is \emph{not only} because they are doubly parametrized unlike many others---i.e., for \emph{high} attacks, not only the majority of words is attacked but also the majority of characters within a word---since they are also effective at \emph{low} attack levels. We partly attribute their success to the fact that they cause a high out-of-vocabulary rate for \emph{RoBERTa} and %possibly even [many segmentations?]
tend to increase the number of input tokens, as they cause \emph{RoBERTa} to segment the input at unknown characters. This may lead to the number of input tokens exceeding \emph{RoBERTa}'s builti-in max token size, leading to cutting off the ending of the sentence.  
}

\begin{figure}
    \centering
    \includegraphics[page=12, width=\linewidth]{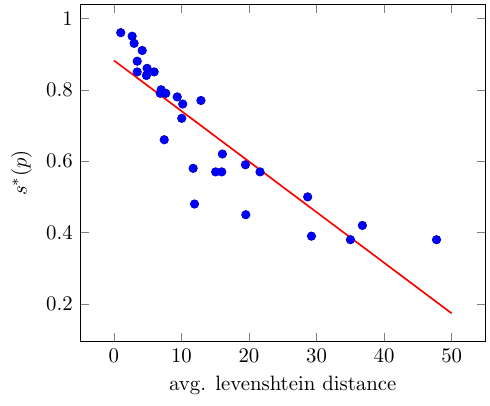}
    \caption{Relation between the amount of text perturbed (measured in edit distance) in a test data set and $s^*(p)$, the performance decrease a model suffers.}
    \label{fig:relation}
\end{figure}

In Table~\ref{tab:attacks_ranking} (appendix), %a ranking for the 
attacks %is given 
are ranked (for \emph{high} attack level) %We rank them 
by the performance degradation caused to the model for each individual task. %\todo[inline]{SE: how exactly did you calculate this number? YB: durchschnitt ueber alle robustness improvements: für 1-1 nur die auf der diagonalen, loo eben alle in der jeweiligen Spalte, gleiches für PLM \& PP (aber die sollten ja sowieso raus)} 
%In addition we also report the readability annotated by humans. 
%Therefore we asked 5 non-native English speakers to rank 10 sentences (see Appendix~\ref{tab:human_readable}) by their readability. 
%Each of these sentences was perturbed by a different attacker beforehand. 
%It is clear to us that no definite statement can be made from this, but we would like to put the machine performances in some relation to human understanding. 
%Not surprisingly and as above mentioned 
In line with our previous discussion, the visual and the intrude attackers are always the both best performing, 
followed by full-shuffle (which we deemed as unrealistic as it would also destroy human perception abilities). 
Figure \ref{fig:relation} shows the relationship between the amount of text perturbed in a test dataset and the performance deterioration a model suffers. This shows a clear (linear) trend and indicates that a successful attacker most importantly needs to attack many characters of a text to be effective, despite all individual qualitative differences between the attackers discussed above.   

{
In Table~\ref{tab:defense_ranking}, a ranking of defense strategies is given. 1-1 ($\alpha,\alpha$) performs best, but is unrealistic. LOO is a robust alternative for unknown new attacks. The effectiveness of LOO as  defense is also a further justification for designing multiple attack models. 
%Except for intrude this applies equally to humans. 
%For visual that was unexpected, because according to \citet{Eger2019} we assumed it to be better readable by humans. 
%We asked the experimentees about their reasons for this particular decision and found out that the main reason was that the visual example contained multiple unknown characters that had to be deciphered. 
%The phonetic attack yields the worst performance for all tasks. This also corresponds to the human annotation.
}

\begin{table}
    \centering
    {\footnotesize
    \begin{tabular}{c | lr|lr|lr}\toprule
        \thead{Rank} & \multicolumn{2}{|c|}{\thead{POS}} & \multicolumn{2}{|c|}{\thead{NLI}} & \multicolumn{2}{|c}{\thead{TC}} \\ \midrule
        %\rowcolor{gold}     
        1 & 1-1 ($\alpha,\alpha$) & 16 & 1-1 & 20  & 1-1  & 12 \\\hline
        %\rowcolor{silver}   
        2 & LOO  & 4  & LOO & 10  & LOO  & 9\\ \hline
        %\rowcolor{bronze}   
        3 & 1-1 ($\alpha$,$\beta$)  & 1  & 1-1  & 3  & 1-1  & 7\\ %\hline
        %\rowcolor{bronze}   3 & PP      & +1\%  & PP     & +1\% %  & PLM     & +8\%   \\\hline
        %                    4 & PLM     & -1\%  & PLM    & 0\%    & PP      & +6\%\\
        \bottomrule
    \end{tabular}
    \caption{Different defense approaches ranked by the average robustness improvement over all attackers. Improvement in percentage points (pp; rounded).}
    \label{tab:defense_ranking}
    }
\end{table}

\section{Conclusion}
\label{sec:conclusion}
We provided the first large-scale catalogue for low-level adversarial attacks, providing a new simple benchmark for testing real-world robustness of future deep learning models. We further showed that one of the currently most successful deep learning paradigms, \emph{RoBERTa}, is not robust to our benchmark, sometimes suffering catastrophic failure. While many of our errors could probably be addressed by placing a correction layer in front of \emph{RoBERTa} \citep{choudhury2007investigation,pruthi-etal-2019-combating}, we believe that our findings shed further light on the differences between human and machine text processing, which deep models eventually will have to \emph{innately} overcome for true AI to become a viable prospect.

\section*{Acknowledgments}
We thank the anonymous reviewers for their useful comments and suggestions. Steffen Eger has been funded by the HMWK (Hessisches Ministerium für Wissenschaft und Kunst) as part of structural location promotion for TU Darmstadt in the context of the Hessian excellence cluster initiative ``Content Analytics for the Social Good'' (CA-SG). 

\bibliography{aacl-ijcnlp2020}
%\bibliography{anthology,aacl-ijcnlp2020}
\bibliographystyle{acl_natbib}

%\iffalse
\clearpage

\appendix

\section{Appendices}
\label{sec:appendix}
\subsection{Phonetic and visual embeddings}
\paragraph{Phonetic Word Embeddings.}\label{par:phonetic_word_embeddings}
To induce phonetic word embeddings, we 
adopt %the architecture of \textit{InferSent} first introduced by \citet{Conneau2018}. 
the Siamese network of InferSent \citep{Conneau2018}. InferSent was originally designed to induce vector representations for two sentences from which their entailment relation was inferred. We adapt InferSent to encode two words so that their phonological similarity can be inferred: \textit{identical}, \textit{very similar}, \textit{similar} and \textit{different}. We use the BiLSTM max-pooling approach from the original InferSent paper, where we set the induced phonetic embeddings size to 100.

We build our own dataset for phonetic similarity by leveraging data from different sources. %, because no such data was available yet. 
%A good starting point were pronunciation dictionaries like \textit{Combilex} \citep{combilex} which assign each word its associated phoneme. 
Initially, we use Combilex \citep{Richmond2010}, which gives phonetic representations for standard (American) English words. 
%To obtain a comparable measurement of phonetic similarity, 
We calculate the normalized edit distance between the phonemes of each word pair %and divided it by the length of the shorter phoneme to normalize it:
to determine the phonetic similarity of two words: 
\begin{equation}\label{eq:phonetic_sim}
    \text{sim}_{\text{ph}}(\pi_1,\pi_2)=1-\frac{d(\pi_1,\pi_2)}{\min(|\pi_1|,|\pi_2|)}
\end{equation}
%\todo{\sout{SE: I think usually the max is taken for normalized edit distance ...} YB: legitimate concern - Sadly I cannot fix this... models are already trained, but sure the edit distance can be larger than min(w1,w2)}
%\todo{\sout{SE: In Eq (1), you show a distance, but below you talk about similarity.} YB: 1 - difference => similarity}
where $\pi_i$ are % \in\{1,2\}$ are %the phonemes 
phonetic sequences for underlying words 
and $d$ is the edit-distance. %The resulting value represents the phonetic similarity percentage between the two input words. 
%Afterwards, %these words were clustered 
We then map the words into 4 different classes: \textit{identical} ($\text{sim}_{\text{ph}}=0$), \textit{very similar} ($0<\text{sim}_{\text{ph}}<0.1$), \textit{similar} ($0.1<\text{sim}_{\text{ph}}<0.3$) and \textit{different} ($0.3<\text{sim}_{\text{ph}}$). %to build our training bins. 
%The resulting class distribution is shown in Figure~\ref{fig:data_distribution}.\todo{SE: you take all words from InferSent? How many are these? How is the data constructed?}
%As expected there are few samples in \textit{very similar} and a lot of samples in the \textit{different} bin. 
To keep the training data for each class more balanced, we added handcrafted and crawled samples, %by hand, 
e.g., homophones. %(references are given in the appendix). 
%\todo{SE: something is missing here?} 
%There was also a lack of 
We also wanted to include ``internet slang'' style phonetic replacements like in Table~\ref{tab:phoneticexamples}. We therefore crawled
%\footnote{\fullcite{Possle_abbrev}\\\fullcite{Beal_messageAbbreviations}} 
them %manually 
and added them to the bins \textit{identical} and \textit{very similar} %with best knowledge and certainty. 
based upon manual inspection. 
%After adding new data to the emptier bins there were finally enough data to slice it into four equal bins with 5k samples each. 
Overall, we compiled 5k examples for each of our four labels. 
The \textit{similar} and \textit{different} bins consist only of data from Combilex, whereas the \textit{identical} and \textit{very similar} bin contains 1.3k samples from Combilex and 3.7k crawled samples. %\todo{SE: where did you crawl them from?}
References for crawled sites are given in \ref{sec:homophone_resources}.

\begin{table}
    \centering
    \begin{tabular}{l|cc}
        \toprule
            \textbf{Homophone} & byte & bite \\ %\midrule
            \textbf{Abbreviation} & I love you too! & I luv U 2! \\
        \bottomrule
    \end{tabular}
    \caption{Example of a homophone and a typical ``internet slang" abbreviation.}
    \label{tab:phoneticexamples}
\end{table}

\begin{figure}[ht]
    \centering
    \includegraphics[width=\linewidth,page=8]{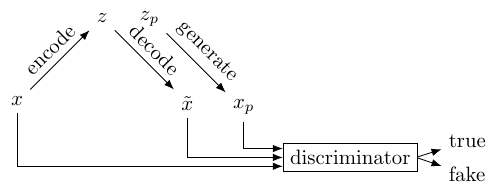}
    \caption{Schematic representation of \gls{vaegan} taken and adapted from \citet{Larsen2016}. $z$ can be decomposed as $z=\mu+\sigma$ and is used to sample $z_p=\mu+\sigma\epsilon$ where $\epsilon$ is noise defined as $\epsilon \sim\ \mathcal{N}(0,I)$}
    \label{fig:vaegan_arch}
\end{figure}

\paragraph{Visual Embeddings.} 
The model reduces the dimension of input $x$, e.g., an image, by applying multiple convolutional steps in the \textit{encoder} to compute the latent representation $z$ of $x$. Afterwards, it reconstructs the original input $x$ in the \textit{decoder} by applying multiple deconvolutional steps to $z$. This reconstructed version of $x$ is called $\tilde{x}$. Additionally, a second input $z_p$ %a second input 
sampled from $\mathcal{N}(0,I)$ is inserted into the \textit{generator} to obtain $x_p$. \textit{Decoder} and \textit{generator} perform the same task on different inputs; they can be considered as identical and therefore share their parameters.
% Although the \textit{decoder} and the \textit{generator} execute the same task, they are considered as identical and therefore share their parameters.
%\todo{\sout{SE: although the do the same task, they are considered identical?}}
%The last remaining component is the \textit{discriminator}. It 
The \textit{discriminator} 
takes $x,\tilde{x}$ and $x_p$ as inputs and discriminates which input is a real training sample and which is a fake. Figure \ref{fig:vaegan_arch} illustrates the working of the architecture. 

\begin{figure}[ht]
    \centering
    \includegraphics[width=\linewidth]{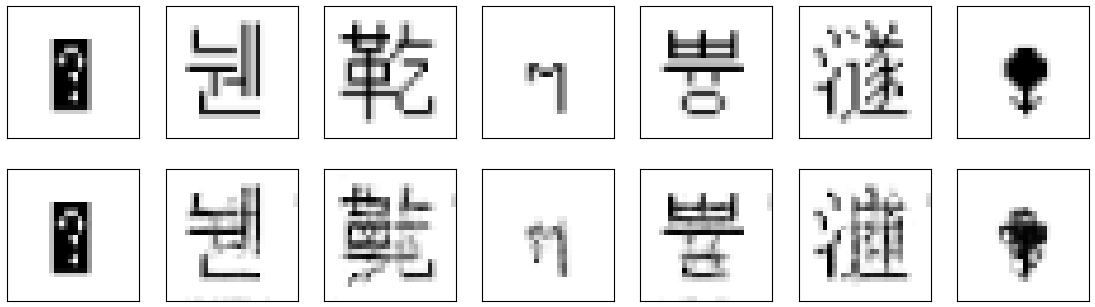}
    \caption{Reconstruction of images after being compressed to its latent representation and decompressed back to the original data distribution.}
    \label{fig:vaegan_reconstruction}
\end{figure}

%The peculiarity of generative models like this \gls{vaegan} is that they improve the performance of their components by having them compete against each other. 
%While the generator is improving its performance of generating samples, which match the original data distribution, out of noise, the discriminator gets better at distinguishing between real and fake images. 
%In addition 
%A specific property of the \gls{vaegan} is that it also improves the ability of the encoder/decoder to properly compress/decompress the samples/latent representations in this process.

\begin{figure*}
    \centering
    \includegraphics[width=\linewidth]{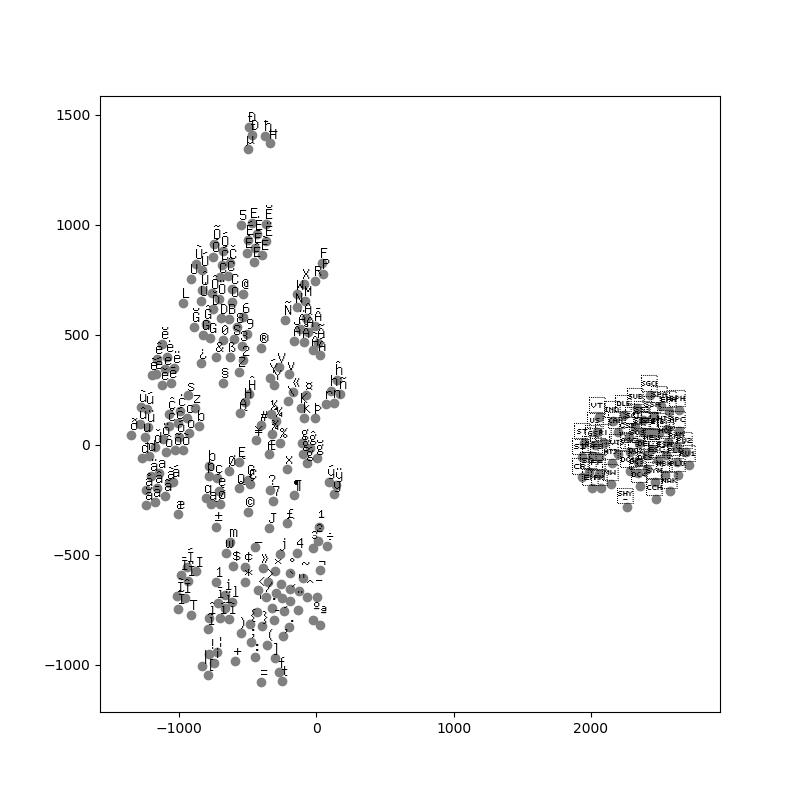}
    \caption{tSNE plot of our character embedding space. As can be seen similar looking characters are clustered.}
    \label{fig:vaegan_tsne}
\end{figure*}

Figure~\ref{fig:vaegan_tsne} %in the appendix 
gives an impression of the encoded visual similarity.

\begin{table*}
    \centering
    { %\footnotesize
    \begin{tabular}{c|l|l|l|l}\toprule
        \thead{Ranking} & \thead{POS} & \thead{NLI} & \thead{TC} %& \thead{Human} 
        \\\hline
        1    & Intrude           & Visual \& Intrude & Visual        %& Visual \& Keyboard-Typo 
        \\\hline
        2  & Visual            & -                 & Intrude       %& - 
        \\\hline
        3  & Full-Shuffle      & Full-Shuffle      & Full-Shuffle  %& Full-Shuffle 
        \\\hline
        4                   & Keyboard-Typo     & Disemvowel        & Disemvowel    %& Natural-Noise 
        \\\hline
        5                   & Disemvowel        & Keyboard-Typo     & Inner-Shuffle %& Disemvoweling 
        \\\hline
        6                   & Natural-Noise     & Inner-Shuffle     & Truncate      %& Intrude \& Inner-Shuffle 
        \\\hline
        7                   & Inner-Shuffle     & Natural-Noise     & Keyboard-Typo %& - 
        \\\hline
        8                   & Truncate          & Segment           & Natural-Noise %& Truncate 
        \\\hline
        9                   & Phonetic          & Truncate          & Segment       %& Segment 
        \\\hline
        10                  & -                 & Phonetic          & Phonetic      %& Phonetic  
        \\\bottomrule
    \end{tabular}
    \caption{Ranking on harmfulness of the attackers on \acrshort{pos}, \acrshort{nli}, \acrshort{tc} %and humans 
    on attack level \textit{high}.}
    \label{tab:attacks_ranking}
}
    
\end{table*}

\subsection{Homophone resources}
\label{sec:homophone_resources}
List of used resources to gather homophones.
%\todo[color=green]{YB: check if referenced as expected they are also properly formatted for citing in the website\_references bib file and can be included via cite command.}

\begin{itemize}
    \item \url{https://7esl.com/homonyms/}
    \item \url{https://www.englishclub.com/pronunciation/homophones-list.html}
    \item \url{https://www.thoughtco.com/homonyms-homophones-and-homographs-a-b-1692660}
    \item \url{http://www.singularis.ltd.uk/bifroest/misc/homophones-list.html}
    \item \url{https://web.archive.org/web/20160825095711/}
    \item \url{http://people.sc.fsu.edu/~jburkardt/fun/wordplay/multinyms.html}
    \item \url{http://homophonelist.com/homophones-list/}
    \item \url{https://web.archive.org/web/20160825095711/}
    \item \url{http://homophonelist.com/homophones-list/}
    \item \url{https://www.webopedia.com/quick\_ref/textmessageabbreviations.asp}
    \item \url{https://www.smart-words.org/abbreviations/text.html}
    \item \url{https://en.wiktionary.org/wiki/Appendix:English\_dialect-independent\_homophones}
    \item \url{https://en.wiktionary.org/wiki/Appendix:English\_dialect-dependent\_homophones}
\end{itemize}

\subsection{Detailed Result Tables}
Hyperparameters of our models can be found in the github accompanying the publication (\url{https://github.com/yannikbenz/zeroe}). The following tables give detailed results of our experiments.
\label{sec:extended_result_tables}

%-------------------------------
% Attack against clean model performance table
\begin{table*}[ht]
\centering
\begin{tabular}{@{\extracolsep{4pt}}lcccc@{}} \\ \toprule
\multirow{2}{*}{\thead{Attack}} & \multirow{2}{*}{\thead{Mode}} & \multicolumn{2}{c}{\thead{Accuracy}} & \thead{AUCROC} \\
\cline{3-4} \cline{5-5}
& & \thead{POS} & \thead{NLI} & \thead{TC} \\ \midrule
None & - & 96.65 & 90.41 & 0.93 \\ \midrule
\multirow{3}{*}{Full-Swap}
    & low   & 82.14 & 70.35 & 0.90  \\ \cline{2-5}
    & mid   & 58.14 & 45.70 & 0.83  \\ \cline{2-5}
    & high  & 40.47 & 38.35 & 0.74  \\ \midrule
\multirow{3}{*}{Inner-Swap}
    & low   & 85.96 & 67.70 & 0.90 \\\cline{2-5}
    & mid   & 70.53 & 53.35 & 0.83 \\\cline{2-5}
    & high  & 67.95 & 51.55 & 0.82 \\\midrule
\multirow{3}{*}{Intrude} 
    & low   & 81.42 & 75.97 & 0.91 \\\cline{2-5}
    & mid   & 46.91 & 52.25 & 0.85 \\\cline{2-5}
    & high  & 18.15 & 34.70 & 0.66 \\\midrule
\multirow{3}{*}{Disemvowel} 
    & low   &   85.24   & 72.24 & 0.91 \\\cline{2-5}
    & mid   &   61.50   & 51.62 & 0.86 \\\cline{2-5}
    & high  &   44.69   & 41.00 & 0.79 \\\midrule
\multirow{3}{*}{Truncate} 
    & low   &   88.57   & 79.83 & 0.90 \\\cline{2-5}
    & mid   &   77,40   & 72.87 & 0.84 \\\cline{2-5}
    & high  &   75,11   & 72.02 & 0.83 \\\midrule
\multirow{3}{*}{Segment} 
    & low   &   -   & 86.08 & 0.93 \\\cline{2-5}
    & mid   &   -   & 77.53 & 0.92 \\\cline{2-5}
    & high  &   -   & 69.14 & 0.91 \\\midrule
\multirow{3}{*}{Keyboard-Typo} 
    & low   &   85.06   & 76.93 & 0.92 \\\cline{2-5}
    & mid   &   62.41   & 60.21 & 0.88 \\\cline{2-5}
    & high  &   40.99   & 44.16 & 0.84 \\\midrule
\multirow{3}{*}{Natural Noise} 
    & low   &   85.34   & 78.43 & 0.92 \\\cline{2-5}
    & mid   &   65.36   & 65.60 & 0.91 \\\cline{2-5}
    & high  &   50.06   & 56.31 & 0.90 \\\midrule
\multirow{3}{*}{Phonetic} 
    & low   &   90.62   & 87.40 & 0.93 \\\cline{2-5}
    & mid   &   89.09   & 84.75 & 0.92 \\\cline{2-5}
    & high  &   88.95   & 82.80 & 0.91 \\\midrule
\multirow{3}{*}{Visual}
    & low   &   80.52   & 53.07 & 0.86 \\\cline{2-5}
    & mid   &   48.14   & 35.26 & 0.64 \\\cline{2-5}
    & high  &   22.44   & 34.37 & 0.48 \\\bottomrule
\end{tabular}
\caption{Attacks against unshielded model.}
\label{tab:attack-performance}
\end{table*}
%-------------------------------

%-------------------------------
% AT leave-one-out POS/NLI & TC
\begin{table*}[ht]
    \centering
    \addtolength{\leftskip} {-2cm}
    \addtolength{\rightskip}{-2cm}
    \begin{tabular}{c|c|cccccccccc|KKR}
        \toprule
        \diagbox{Test}{Train} & level & FS & IS & INT & DIS & TRUN & SEG & KEY & NAT & PH & VIS & \multicolumn{1}{c}{POS} & \multicolumn{1}{c}{NLI} & \multicolumn{1}{c}{TC} \\ \midrule
        \multirow{4}{*}{FS} & none & \multirow{4}{*}{{\color{red} \huge \xmark}} & \multirow{4}{*}{{\huge \cmark}} & \multirow{4}{*}{{\huge \cmark}} & \multirow{4}{*}{{\huge \cmark}} & \multirow{4}{*}{{\huge \cmark}} & \multirow{4}{*}{{\huge \cmark}} & \multirow{4}{*}{{\huge \cmark}} & \multirow{4}{*}{{\huge \cmark}} & \multirow{4}{*}{{\huge \cmark}}  & \multirow{4}{*}{{\huge \cmark}} & 95.57 & 89.56 & 0.97 \\
                            & low  & & & & & & & & & & & 84.49 & 73.05 & 0.95 \\
                            & mid  & & & & & & & & & & & 63.48 & 57.54 & 0.90 \\
                            & high & & & & & & & & & & & 45.72 & 51.73 & 0.86 \\ \midrule
        \multirow{4}{*}{IS} & none & \multirow{4}{*}{{\huge \cmark}} & \multirow{4}{*}{{\color{red} \huge \xmark}} & \multirow{4}{*}{{\huge \cmark}} & \multirow{4}{*}{{\huge \cmark}} & \multirow{4}{*}{{\huge \cmark}} & \multirow{4}{*}{{\huge \cmark}} & \multirow{4}{*}{{\huge \cmark}} & \multirow{4}{*}{{\huge \cmark}} & \multirow{4}{*}{{\huge \cmark}}  & \multirow{4}{*}{{\huge \cmark}} & 95.66 & 88.94 & 0.96 \\
                            & low  & & & & & & & & & & & 88.29 & 75.51 & 0.94 \\
                            & mid & & & & & & & & & & & 75.90 & 69.07 & 0.91 \\
                            & high & & & & & & & & & & & 73.68 & 68.54 & 0.90 \\ \midrule
        \multirow{4}{*}{INT} & none & \multirow{4}{*}{{\huge \cmark}} & \multirow{4}{*}{{\huge \cmark}} & \multirow{4}{*}{{\color{red} \huge \xmark}} & \multirow{4}{*}{{\huge \cmark}} & \multirow{4}{*}{{\huge \cmark}} & \multirow{4}{*}{{\huge \cmark}} & \multirow{4}{*}{{\huge \cmark}} & \multirow{4}{*}{{\huge \cmark}} & \multirow{4}{*}{{\huge \cmark}}  & \multirow{4}{*}{{\huge \cmark}} & 95.65 & 88.90 & 0.96 \\
                            & low  & & & & & & & & & & & 87.54 & 84.27 & 0.95 \\
                            & mid & & & & & & & & & & & 57.58 & 74.92 & 0.93 \\
                            & high & & & & & & & & & & & 19.44 & 61.07 & 0.84 \\ \midrule
        \multirow{4}{*}{DIS} & none & \multirow{4}{*}{{\huge \cmark}} & \multirow{4}{*}{{\huge \cmark}} & \multirow{4}{*}{{\huge \cmark}} & \multirow{4}{*}{{\color{red} \huge \xmark}} & \multirow{4}{*}{{\huge \cmark}} & \multirow{4}{*}{{\huge \cmark}} & \multirow{4}{*}{{\huge \cmark}} & \multirow{4}{*}{{\huge \cmark}} & \multirow{4}{*}{{\huge \cmark}}  & \multirow{4}{*}{{\huge \cmark}} & 95.69 & 89.42 & 0.96 \\
                            & low  & & & & & & & & & & & 86.00 & 80.00 & 0.94 \\
                            & mid & & & & & & & & & & & 64.39 & 70.98 & 0.91 \\
                            & high & & & & & & & & & & & 48.65 & 66.60 & 0.89 \\ \midrule
        \multirow{4}{*}{TRUN} & none & \multirow{4}{*}{{\huge \cmark}} & \multirow{4}{*}{{\huge \cmark}} & \multirow{4}{*}{{\huge \cmark}} & \multirow{4}{*}{{\huge \cmark}} & \multirow{4}{*}{{\color{red} \huge \xmark}} & \multirow{4}{*}{{\huge \cmark}} & \multirow{4}{*}{{\huge \cmark}} & \multirow{4}{*}{{\huge \cmark}} & \multirow{4}{*}{{\huge \cmark}}  & \multirow{4}{*}{{\huge \cmark}} & 95.49 & 89.17 & 0.96 \\
                            & low  & & & & & & & & & & & 89.98 & 84.55 & 0.84 \\
                            & mid & & & & & & & & & & & 81.23 & 81.97 & 0.83 \\
                            & high & & & & & & & & & & & 79.38 & 81.62 & 0.82 \\ \midrule
        \multirow{4}{*}{SEG} & none & \multirow{4}{*}{{\huge \cmark}} & \multirow{4}{*}{{\huge \cmark}} & \multirow{4}{*}{{\huge \cmark}} & \multirow{4}{*}{{\huge \cmark}} & \multirow{4}{*}{{\huge \cmark}} & \multirow{4}{*}{{\color{red} \huge \xmark}} & \multirow{4}{*}{{\huge \cmark}} & \multirow{4}{*}{{\huge \cmark}} & \multirow{4}{*}{{\huge \cmark}}  & \multirow{4}{*}{{\huge \cmark}} & \multicolumn{1}{c}{-} & 89.02 & 0.96 \\
                            & low  & & & & & & & & & & & \multicolumn{1}{c}{-} & 85.38 & 0.96 \\
                            & mid & & & & & & & & & & & \multicolumn{1}{c}{-} & 76.92 & 0.95 \\
                            & high & & & & & & & & & & & \multicolumn{1}{c}{-} & 62.83 & 0.95 \\ \midrule
        \multirow{4}{*}{KEY} & none & \multirow{4}{*}{{\huge \cmark}} & \multirow{4}{*}{{\huge \cmark}} & \multirow{4}{*}{{\huge \cmark}} & \multirow{4}{*}{{\huge \cmark}} & \multirow{4}{*}{{\huge \cmark}} & \multirow{4}{*}{{\huge \cmark}} & \multirow{4}{*}{{\color{red} \huge \xmark}} & \multirow{4}{*}{{\huge \cmark}} & \multirow{4}{*}{{\huge \cmark}}  & \multirow{4}{*}{{\huge \cmark}} & 95.61 & 88.80 & 0.96 \\
                            & low  & & & & & & & & & & & 87.71 & 80.47 & 0.95 \\
                            & mid & & & & & & & & & & & 68.64 & 70.69 & 0.94 \\
                            & high & & & & & & & & & & & 46.51 & 61.83 & 0.92 \\ \midrule
        \multirow{4}{*}{NAT} & none & \multirow{4}{*}{{\huge \cmark}} & \multirow{4}{*}{{\huge \cmark}} & \multirow{4}{*}{{\huge \cmark}} & \multirow{4}{*}{{\huge \cmark}} & \multirow{4}{*}{{\huge \cmark}} & \multirow{4}{*}{{\huge \cmark}} & \multirow{4}{*}{{\huge \cmark}} & \multirow{4}{*}{{\color{red} \huge \xmark}} & \multirow{4}{*}{{\huge \cmark}}  & \multirow{4}{*}{{\huge \cmark}} & 95.72 & 88.67 & 0.96 \\
                            & low  & & & & & & & & & & & 88.17 & 81.27 & 0.96 \\
                            & mid & & & & & & & & & & & 72.30 & 73.05 & 0.96 \\
                            & high & & & & & & & & & & & 56.78 & 67.40 & 0.96 \\ \midrule
        \multirow{4}{*}{PH} & none & \multirow{4}{*}{{\huge \cmark}} & \multirow{4}{*}{{\huge \cmark}} & \multirow{4}{*}{{\huge \cmark}} & \multirow{4}{*}{{\huge \cmark}} & \multirow{4}{*}{{\huge \cmark}} & \multirow{4}{*}{{\huge \cmark}} & \multirow{4}{*}{{\huge \cmark}} & \multirow{4}{*}{{\huge \cmark}} & \multirow{4}{*}{{\color{red} \huge \xmark}} & \multirow{4}{*}{{\huge \cmark}} & 95.30 & 88.95 & 0.96 \\
                            & low  & & & & & & & & & & & 89.74 & 87.54 & 0.96 \\
                            & mid & & & & & & & & & & & 87.82 & 86.27 & 0.95 \\
                            & high & & & & & & & & & & & 87.72 & 85.34 & 0.95 \\ \midrule
        \multirow{4}{*}{VIS} & none & \multirow{4}{*}{{\huge \cmark}} & \multirow{4}{*}{{\huge \cmark}} & \multirow{4}{*}{{\huge \cmark}} & \multirow{4}{*}{{\huge \cmark}} & \multirow{4}{*}{{\huge \cmark}} & \multirow{4}{*}{{\huge \cmark}} & \multirow{4}{*}{{\huge \cmark}} & \multirow{4}{*}{{\huge \cmark}} & \multirow{4}{*}{{\huge \cmark}}  & \multirow{4}{*}{{\color{red} \huge \xmark}} & 95.72 & 89.02 & 0.96 \\
                            & low  & & & & & & & & & & & 85.18 & 70.77 & 0.93 \\
                            & mid & & & & & & & & & & & 58.94 & 48.80 & 0.85 \\
                            & high & & & & & & & & & & & 24.99 & 40.22 & 0.75 \\
        \bottomrule
    \end{tabular}
    \caption{Adversarial training: leave-one-out.}
    \label{tab:adversarial-training-leave-one-out}
\end{table*}
%-------------------------------

%-------------------------------
% AT 1-1 POS
\begin{table*}[ht]
    \centering
    \begin{tabular}{c|c|KKKKKKKKKK}
        \toprule
        \diagbox{Test}{Train} & level & \multicolumn{1}{c}{FS} & \multicolumn{1}{c}{IS} & \multicolumn{1}{c}{INT} & \multicolumn{1}{c}{DIS} & \multicolumn{1}{c}{TRUN} & \multicolumn{1}{c}{KEY} & \multicolumn{1}{c}{NAT} & \multicolumn{1}{c}{PH} & \multicolumn{1}{c}{VIS} \\ \midrule
        Clean & - & 95.17 & 95.18 & 95.04 & 95.44 & 95.48 & 95.40 & 95.40 & 96.06 & 95.66 \\ \midrule
        \multirow{3}{*}{FS} 
        & low   & 88.49 & 81.68 & 80.01 & 82.63 & 81.85 & 81.45 & 80.93 & 82.16 & 83.10\\
        & mid   & 75.73 & 58.04 & 54.04 & 62.17 & 59.95 & 57.69 & 59.62 & 56.91 & 60.38\\
        & high  & 62.27 & 40.95 & 36.11 & 48.06 & 44.84 & 40.58 & 44.40 & 38.79 & 42.44\\ %\midrule
        \multirow{3}{*}{IS} 
        & low   & 87.97 & 91.98 & 86.06 & 86.68 & 85.65 & 87.09 & 85.63 & 86.10 & 87.70\\
        & mid   & 74.23 & 87.24 & 71.42 & 74.77 & 72.93 & 74.06 & 70.94 & 70.61 & 74.61\\
        & high  & 71.72 & 86.24 & 69.20 & 73.20 & 70.47 & 71.84 & 68.54 & 67.67 & 72.73\\ %\midrule
        \multirow{3}{*}{INT} 
        & low   & 79.23 & 80.38 & 92.72 & 81.82 & 82.05 & 86.88 & 82.27 & 79.26 & 86.72\\
        & mid   & 39.12 & 42.02 & 88.41 & 57.68 & 54.17 & 59.29 & 53.62 & 44.37 & 60.84\\
        & high  & 10.46 & 11.93 & 80.96 & 27.95 & 24.74 & 29.55 & 47.71 & 15.88 & 22.45\\ %\midrule
        \multirow{3}{*}{DIS} 
        & low   & 82.29 & 82.45 & 84.68 & 93.31 & 85.91 & 85.43 & 85.72 & 84.27 & 85.59\\
        & mid   & 56.75 & 60.08 & 62.94 & 89.23 & 67.55 & 66.68 & 67.18 & 60.33 & 64.77\\
        & high  & 40.30 & 45.11 & 47.45 & 86.13 & 53.89 & 53.37 & 54.08 & 42.87 & 49.31\\ %\midrule
        \multirow{3}{*}{TRUN} 
        & low   & 88.17 & 87.50 & 88.34 & 88.97 & 94.58 & 89.45 & 90.03 & 88.63 & 88.80\\
        & mid   & 77.18 & 76.02 & 78.33 & 79.58 & 93.27 & 80.58 & 82.33 & 76.79 & 78.98\\
        & high  & 74.91 & 73.84 & 76.28 & 77.90 & 92.99 & 78.64 & 80.84 & 74.33 & 76.83\\ %\midrule
        \multirow{3}{*}{KEY} 
        & low   & 83.70 & 83.16 & 85.17 & 85.36 & 85.30 & 90.11 & 85.97 & 84.11 & 86.99\\
        & mid   & 60.52 & 60.99 & 63.81 & 65.23 & 64.07 & 73.45 & 67.06 & 60.62 & 66.53\\
        & high  & 36.92 & 38.50 & 44.95 & 44.25 & 45.64 & 62.99 & 47.71 & 37.72 & 46.93\\ %\midrule
        \multirow{3}{*}{NAT} 
        & low   & 83.92 & 82.57 & 85.21 & 85.13 & 84.74 & 85.83 & 88.52 & 84.41 & 86.48\\
        & mid   & 65.05 & 62.50 & 65.83 & 68.13 & 66.39 & 67.97 & 80.98 & 64.59 & 67.51\\
        & high  & 47.83 & 47.01 & 52.39 & 53.52 & 54.10 & 54.26 & 66.76 & 48.79 & 53.25\\ %\midrule
        \multirow{3}{*}{PH} 
        & low   & 88.82 & 88.66 & 88.89 & 89.40 & 89.21 & 88.86 & 89.81 & 94.89 & 89.47\\
        & mid   & 86.68 & 86.91 & 87.12 & 87.64 & 87.36 & 86.93 & 88.07 & 94.39 & 87.67\\
        & high  & 86.74 & 86.79 & 87.00 & 87.51 & 87.26 & 86.85 & 88.00 & 94.36 & 87.56\\ %\midrule
        \multirow{3}{*}{VIS} 
        & low   & 82.07 & 82.13 & 83.10 & 82.64 & 82.61 & 84.21 & 82.26 & 81.20 & 87.95\\
        & mid   & 53.15 & 54.79 & 56.05 & 57.27 & 55.55 & 58.06 & 53.23 & 49.32 & 66.58\\
        & high  & 23.90 & 26.06 & 22.19 & 30.31 & 29.04 & 30.43 & 26.25 & 23.66 & 31.38\\
        \bottomrule
    \end{tabular}
    \caption{Part-of-Speech tagging adversarial training: 1-1.}
    \label{tab:1-1-adv-training_pos}
\end{table*}
%-------------------------------

%-------------------------------
% AT 1-1 NLI
\begin{table*}[ht]
    \centering
    \begin{tabular}{c|c|KcKKKKKKKc}
        \toprule
        \diagbox{Test}{Train} & level & \multicolumn{1}{c}{FS} & \multicolumn{1}{c}{IS} & \multicolumn{1}{c}{INT} & \multicolumn{1}{c}{DIS} & \multicolumn{1}{c}{TRUN} & \multicolumn{1}{c}{SEG} & \multicolumn{1}{c}{KEY} & \multicolumn{1}{c}{NAT} & \multicolumn{1}{c}{PH} & \multicolumn{1}{c}{VIS} \\ \midrule
        Clean & - & 87.54 & - & 88.29 & 88.59 & 88.91 & 89.90 & 89.17 & 89.12 & 90.24 & - \\ \midrule
        \multirow{3}{*}{FS} 
            % level& FS             & IS    & INT   & DIS   & TRUN  & SEG   & KEY   & NAT   & PH    & VIS
            & low  & 83.04 & -     & 64.65 & 63.28 & 60.38 & 62.67 & 69.46 & 68.46 & 66.87 & - \\
            & mid  & 78.58 & -     & 47.62 & 48.15 & 42.21 & 46.05 & 52.63 & 50.81 & 48.11 & - \\
            & high & 76.96 & -     & 42.75 & 44.52 & 39.16 & 41.55 & 47.77 & 46.21 & 41.53 & - \\ %\midrule
        \multirow{3}{*}{IS} 
            % level& FS    &             IS & INT   & DIS   & TRUN  & SEG   & KEY   & NAT   & PH    & VIS
            & low  & 81.31 & -     & 71.32 & 66.40 & 59.81 & 63.26 & 72.23 & 72.25 & 66.40 & - \\
            & mid  & 78.82 & -     & 64.82 & 58.42 & 51.17 & 53.63 & 63.21 & 64.24 & 57.16 & - \\
            & high & 78.08 & -     & 64.06 & 58.25 & 50.86 & 53.53 & 62.72 & 62.89 & 56.61 & - \\ %\midrule
        \multirow{3}{*}{INT} 
            % level& FS    &    IS & INT            & DIS   & TRUN  & SEG   & KEY   & NAT   & PH    & VIS
            & low  & 76.22 & -     & 85.83 & 72.49 & 72.97 & 72.78 & 83.73 & 80.83 & 74.83 & - \\
            & mid  & 58.99 & -     & 82.61 & 53.87 & 51.09 & 48.45 & 69.53 & 62.84 & 51.53 & - \\
            & high & 43.22 & -     & 80.76 & 39.93 & 36.18 & 36.60 & 48.77 & 40.91 & 37.07 & - \\ %\midrule
        \multirow{3}{*}{DIS} 
            % level& FS    & IS    & INT   & DIS            & TRUN  & SEG   & KEY   & NAT   & PH    & VIS
            & low  & 79.14 & -     & 76.27 & 86.56 & 67.51 & 72.52 & 78.96 & 77.77 & 72.65 & - \\
            & mid  & 72.47 & -     & 67.43 & 84.70 & 57.60 & 56.88 & 69.72 & 64.85 & 57.03 & - \\
            & high & 69.45 & -     & 63.90 & 84.16 & 54.50 & 48.25 & 65.29 & 58.44 & 48.92 & - \\ %\midrule
        \multirow{3}{*}{TRUN} 
            % level& FS    & IS    & INT   & DIS   & TRUN           & SEG   & KEY   & NAT   & PH    & VIS
            & low  & 81.63 & -     & 84.31 & 79.66 & 88.15 & 80.02 & 86.35 & 84.79 & 80.46 & - \\
            & mid  & 77.87 & -     & 82.45 & 75.86 & 87.52 & 76.11 & 84.08 & 81.83 & 76.19 & - \\
            & high & 77.25 & -     & 82.46 & 75.10 & 87.44 & 75.79 & 83.80 & 81.76 & 75.80 & - \\ %\midrule
        \multirow{3}{*}{SEG} 
            % level& FS    & IS    & INT   & DIS   & TRUN  & SEG            & KEY   & NAT   & PH    & VIS
            & low  & 82.32 & -     & 84.32 & 83.75 & 84.00 & 89.07 & 86.15 & 85.53 & 85.69 & - \\
            & mid  & 68.85 & -     & 76.41 & 75.84 & 77.33 & 87.54 & 80.28 & 79.37 & 78.14 & - \\
            & high & 50.94 & -     & 64.98 & 68.11 & 71.36 & 86.42 & 73.39 & 73.19 & 71.88 & - \\ %\midrule
        \multirow{3}{*}{KEY} 
            % level& FS    & IS    & INT   & DIS   & TRUN  & SEG   & KEY            & NAT   & PH    & VIS
            & low  & 74.23 & -     & 76.90 & 71.38 & 69.95 & 73.10 & 86.63 & 81.04 & 74.46 & - \\
            & mid  & 57.37 & -     & 62.86 & 55.62 & 54.30 & 58.67 & 82.98 & 70.74 & 56.98 & - \\
            & high & 45.87 & -     & 52.26 & 46.29 & 44.75 & 47.07 & 79.82 & 61.76 & 44.54 & - \\ %\midrule
        \multirow{3}{*}{NAT} 
            % level& FS    & IS    & INT   & DIS   & TRUN  & SEG   & KEY   & NAT            & PH    & VIS
            & low  & 77.87 & -     & 78.32 & 73.62 & 73.50 & 75.51 & 82.39 & 87.67 & 76.47 & - \\
            & mid  & 67.98 & -     & 67.98 & 62.27 & 60.10 & 62.97 & 74.25 & 85.45 & 62.85 & - \\
            & high & 59.73 & -     & 60.81 & 55.16 & 53.18 & 55.41 & 69.52 & 84.06 & 54.89 & - \\ %\midrule
        \multirow{3}{*}{PH} 
            % level& FS    & IS    & INT   & DIS   & TRUN  & SEG   & KEY   & NAT   & PH    & VIS
            & low  & 85.68 & -     & 85.98 & 84.23 & 85.36 & 86.53 & 87.50 & 87.80 & 89.93 & - \\
            & mid  & 84.25 & -     & 84.21 & 80.98 & 82.67 & 84.17 & 85.98 & 86.50 & 89.40 & - \\
            & high & 83.07 & -     & 82.71 & 80.28 & 81.68 & 82.68 & 84.74 & 85.42 & 89.19 & - \\ %\midrule
        \multirow{3}{*}{VIS} 
            % level& FS    & IS    & INT   & DIS   & TRUN  & SEG   & KEY   & NAT   & PH    & VIS
            & low  & 59.79 & -     & 72.33 & 55.74 & 56.09 & 56.91 & 70.65 & 66.32 & 55.34 & - \\
            & mid  & 41.82 & -     & 50.95 & 37.87 & 37.01 & 36.38 & 45.21 & 41.84 & 36.31 & - \\
            & high & 37.42 & -     & 39.08 & 33.81 & 34.26 & 34.51 & 36.04 & 35.37 & 34.10 & - \\
        \bottomrule
    \end{tabular}
    \caption{Natural language inference adversarial training: 1-1.}
    \label{tab:1-1-adv-training-snli}
\end{table*}
%-------------------------------

%-------------------------------
% AT 1-1 TC
\begin{table*}[ht]
    \centering
    \begin{tabular}{c|c|RRRRRRRRRR}
        \toprule
        \diagbox{Test}{Train} & level & \multicolumn{1}{c}{FS} & \multicolumn{1}{c}{IS} & \multicolumn{1}{c}{INT} & \multicolumn{1}{c}{DIS} & \multicolumn{1}{c}{TRUN} & \multicolumn{1}{c}{SEG} & \multicolumn{1}{c}{KEY} & \multicolumn{1}{c}{NAT} & \multicolumn{1}{c}{PH} & \multicolumn{1}{c}{VIS} \\ \midrule
        Clean & - & 0.96 & 0.96 & 0.96 & 0.97 & 0.97 & 0.98 & 0.97 & 0.96 & 0.97 & 0.96 \\ \midrule
        \multirow{3}{*}{FS} 
        % level & FS            & IS   & INT  & DIS  & TRUN & SEG  & KEY  & NAT  & PH   & VIS
        & low   & 0.94 & 0.94 & 0.94 & 0.94 & 0.94 & 0.95 & 0.95 & 0.93 & 0.95 & 0.93 \\
        & mid   & 0.92 & 0.90 & 0.87 & 0.88 & 0.87 & 0.87 & 0.89 & 0.87 & 0.87 & 0.88 \\
        & high  & 0.90 & 0.86 & 0.80 & 0.81 & 0.79 & 0.79 & 0.83 & 0.80 & 0.77 & 0.83 \\ %\midrule
        \multirow{3}{*}{IS} 
        % level & FS   & IS            & INT  & DIS  & TRUN & SEG  & KEY  & NAT  & PH   & VIS
        & low   & 0.94 & 0.95 & 0.94 & 0.94 & 0.94 & 0.95 & 0.94 & 0.93 & 0.94 & 0.93 \\
        & mid   & 0.92 & 0.94 & 0.89 & 0.90 & 0.89 & 0.89 & 0.91 & 0.88 & 0.89 & 0.89 \\
        & high  & 0.91 & 0.94 & 0.88 & 0.90 & 0.88 & 0.88 & 0.90 & 0.87 & 0.88 & 0.89 \\ %\midrule
        \multirow{3}{*}{INT} 
        % level & FS   & IS   & INT           & DIS  & TRUN & SEG  & KEY  & NAT  & PH   & VIS
        & low   & 0.95 & 0.95 & 0.96 & 0.96 & 0.96 & 0.96 & 0.96 & 0.95 & 0.96 & 0.95 \\
        & mid   & 0.92 & 0.92 & 0.95 & 0.92 & 0.92 & 0.91 & 0.94 & 0.90 & 0.91 & 0.91 \\
        & high  & 0.81 & 0.80 & 0.91 & 0.80 & 0.76 & 0.75 & 0.81 & 0.76 & 0.72 & 0.83 \\ %\midrule
        \multirow{3}{*}{DIS} 
        & low   & 0.94 & 0.95 & 0.95 & 0.96 & 0.94 & 0.96 & 0.95 & 0.94 & 0.95 & 0.94 \\
        & mid   & 0.91 & 0.91 & 0.91 & 0.96 & 0.90 & 0.90 & 0.92 & 0.90 & 0.89 & 0.90 \\
        & high  & 0.88 & 0.88 & 0.88 & 0.95 & 0.86 & 0.83 & 0.89 & 0.86 & 0.83 & 0.88 \\ %\midrule
        \multirow{3}{*}{TRUN} 
        & low   & 0.95 & 0.95 & 0.96 & 0.96 & 0.97 & 0.96 & 0.97 & 0.94 & 0.96 & 0.95 \\
        & mid   & 0.94 & 0.94 & 0.95 & 0.94 & 0.97 & 0.95 & 0.96 & 0.93 & 0.95 & 0.94 \\
        & high  & 0.94 & 0.94 & 0.95 & 0.94 & 0.97 & 0.95 & 0.96 & 0.93 & 0.95 & 0.94 \\ %\midrule
        \multirow{3}{*}{SEG} 
        & low   & 0.96 & 0.96 & 0.96 & 0.96 & 0.97 & 0.97 & 0.97 & 0.96 & 0.97 & 0.95 \\
        & mid   & 0.95 & 0.95 & 0.95 & 0.96 & 0.96 & 0.97 & 0.96 & 0.95 & 0.96 & 0.94 \\
        & high  & 0.94 & 0.94 & 0.94 & 0.95 & 0.95 & 0.97 & 0.95 & 0.93 & 0.95 & 0.93 \\ %\midrule
        \multirow{3}{*}{KEY} 
        & low   & 0.95 & 0.95 & 0.95 & 0.95 & 0.96 & 0.96 & 0.96 & 0.95 & 0.96 & 0.95 \\
        & mid   & 0.92 & 0.92 & 0.93 & 0.93 & 0.93 & 0.94 & 0.95 & 0.92 & 0.93 & 0.92 \\
        & high  & 0.88 & 0.88 & 0.91 & 0.89 & 0.90 & 0.89 & 0.95 & 0.89 & 0.87 & 0.90 \\ %\midrule
        \multirow{3}{*}{NAT} 
        & low   & 0.96 & 0.96 & 0.96 & 0.96 & 0.97 & 0.97 & 0.97 & 0.96 & 0.97 & 0.95 \\
        & mid   & 0.95 & 0.95 & 0.96 & 0.96 & 0.96 & 0.97 & 0.97 & 0.96 & 0.96 & 0.95 \\
        & high  & 0.95 & 0.95 & 0.95 & 0.95 & 0.96 & 0.96 & 0.96 & 0.96 & 0.96 & 0.94 \\ %\midrule
        \multirow{3}{*}{PH} 
        & low   & 0.96 & 0.96 & 0.96 & 0.96 & 0.97 & 0.97 & 0.97 & 0.95 & 0.97 & 0.95 \\
        & mid   & 0.95 & 0.95 & 0.95 & 0.96 & 0.96 & 0.97 & 0.96 & 0.95 & 0.97 & 0.95 \\
        & high  & 0.95 & 0.95 & 0.95 & 0.95 & 0.96 & 0.96 & 0.96 & 0.94 & 0.97 & 0.94 \\ %\midrule
        \multirow{3}{*}{VIS} 
        & low   & 0.92 & 0.93 & 0.94 & 0.93 & 0.93 & 0.93 & 0.94 & 0.91 & 0.92 & 0.93 \\
        & mid   & 0.82 & 0.81 & 0.86 & 0.80 & 0.78 & 0.77 & 0.83 & 0.76 & 0.71 & 0.90 \\
        & high  & 0.70 & 0.69 & 0.75 & 0.65 & 0.62 & 0.62 & 0.66 & 0.64 & 0.55 & 0.85 \\
        \bottomrule
    \end{tabular}
    \caption{Toxic comment adversarial training: 1-1.}
    \label{tab:1-1-adv-training_tc}
\end{table*}

\end{document}